\documentclass[10pt]{article}
\usepackage{latexsym,amsmath,amssymb,graphics,amscd}
\usepackage{multirow}
\usepackage{soul}
\usepackage{booktabs}
\usepackage{tabularx}
\usepackage[hang,footnotesize]{caption}
\usepackage[pdftex]{graphicx}
\usepackage{graphicx}
\usepackage{overpic}
\usepackage{color}
\usepackage{cancel}
\usepackage{tikz}
\usetikzlibrary{decorations.pathreplacing}
\usepackage{xr}
\usepackage[showonlyrefs]{mathtools}

\addtocounter{footnote}{1}
\makeatletter
\@addtoreset{table}{bsection}
\def\thetable{\thesection.\@arabic\c@table}
\def\fps@table{h, t}
\@addtoreset{equation}{section}

\makeatother

\newtheorem{theorem}{Theorem}[section]

\newtheorem{lemma}[theorem]{Lemma}
\newtheorem{remark}[theorem]{Remark}
\newtheorem{proposition}[theorem]{Proposition}

\newtheorem{corollary}[theorem]{Corollary}

\newsavebox{\savepar}

\makeatletter

\begin{document}

\title{\textbf{Singular ridge regression with homoscedastic residuals: generalization error with estimated parameters}}
\author{Lyudmila Grigoryeva$^{1}$ and Juan-Pablo Ortega$^{2, 3}$}
\date{}
\maketitle

\begin{abstract}
This paper characterizes the conditional distribution properties of the finite sample ridge regression estimator and uses that result to evaluate total regression and generalization errors that incorporate the inaccuracies committed at the time of parameter estimation. The paper provides explicit formulas for those errors. Unlike other classical references in this setup, our results take place in a fully singular setup that does not assume the existence of a solution for the non-regularized regression problem. In exchange, we invoke a conditional homoscedasticity hypothesis on the regularized regression residuals that is crucial in our developments.
\end{abstract}

\bigskip


\makeatletter
\addtocounter{footnote}{1} \footnotetext{%
Department of Mathematics and Statistics. Universit\"at Konstanz. Box 146. D-78457 Konstanz. Germany. {\texttt{Lyudmila.Grigoryeva@uni-konstanz.de} }}
\addtocounter{footnote}{1} \footnotetext{%
Universit\"at Sankt Gallen. Bodanstrasse 6.
CH-9000 Sankt Gallen. Switzerland. {\texttt{Juan-Pablo.Ortega@unisg.ch}}}
\addtocounter{footnote}{1} \footnotetext{%
Centre National de la Recherche Scientifique (CNRS). France. }
\makeatother

\medskip

\medskip

\medskip

\section{Introduction}

The ridge regression~\cite{tikhonov:regression, tihonov_ridge_1963, hoerl_ridge_1970} has been introduced as a regularization method to estimate the unknown matrix $B$ in the linear regression model
$
\mathbf{y}= B ^\top \mathbf{x} + \boldsymbol{\varepsilon}, 
$
when the covariance matrix of the explanatory variables $\Sigma _{\bf x} := {\rm Cov} ( {\bf x} , {\bf x} ) \in \Bbb S _p $ is singular or poorly conditioned and, therefore, the standard least squares estimator is ill-defined. The ridge regularization method produces an estimate $ \widehat{B} _\lambda $ ($\lambda$ is a regularization strength that we introduce later on)   whose properties are always formulated in terms of the coefficient $B$ that is assumed to solve the least squares problem. Nevertheless, there are many applications in biology, medicine, physics, engineering, or machine learning (see, for instance,~\cite{Geman1992}) in which an underlying linear model with a $B$ that solves the least squares problem does not exist and only a regularized version of it with parameter $B _\lambda $ is available. This is always the case in high-dimensional problems in which the number of covariates $p$ exceeds the sample size $N$ or whenever there is quasi-collinearity among some of the explanatory variables.  

In this paper we place ourselves in that fully singular situation and nevertheless, we manage to establish the conditional distribution properties of the finite sample ridge regression estimator $ \widehat{B} _\lambda  $ directly in terms of $B _\lambda $ without using the least squares coefficient $B$, whose existence is not needed. These results require that certain hypotheses on the conditional homoscedasticity of the regularized regression residuals hold.

The second main contribution of this paper consists of using the properties of this estimator to write down explicit expressions that evaluate the regression (also called training) and the generalization (also called testing) errors committed by a regularized regression model. We recall that the regression or training error is the one committed by the regularized regression model when calculated with the finite sample that has been used to obtain the estimate $\widehat{B} _\lambda $ of the regression coefficient ${B} _\lambda $; for the generalization or testing error we keep $\widehat{B} _\lambda $ and we compute the error committed by the corresponding regression model using another sample that may have different size or even different statistical properties. In both cases our error formulas incorporate the error committed at the time of parameter estimation using the finite training sample. 

The paper is structured in two main sections. The first one is Section~\ref{Multivariate linear ridge regression and ridge estimator} that contains a description of our setup and the hypotheses that we invoke. It also presents the properties of the ridge estimator in the singular framework that we have chosen to work on. Section~\ref{Evaluation of the ridge regression errors: training and testing} contains the results on the evaluation of the training and testing errors. W	e illustrate our developments with various classical examples that give an idea of the scope of our results.  All the proofs of the results and the technical  details are included in the appendices in Section~\ref{Appendices}.

\medskip

\noindent {\bf Notation and conventions:}
Column vectors are denoted by bold lower or upper case  symbol like $\mathbf{v}$ or $\mathbf{V}$. We write $\mathbf{v} ^\top $ to indicate the transpose of $\mathbf{v} $. Given a vector $\mathbf{v} \in \mathbb{R}  ^n $, we denote its entries by $v_i$, with $i \in \left\{ 1, \dots, n
\right\} $; we also write $\mathbf{v}=(v _i)_{i \in \left\{ 1, \dots, n\right\} }$. 
The symbols $\mathbf{i} _n$ and $ \mathbf{0} _n $ stand for the vectors of length $n$ consisting of ones and zeros, respectively.  We denote by $\mathbb{M}_{n ,  m }$ the space of real $n\times m$ matrices with $m, n \in \mathbb{N} $. When $n=m$, we use the symbol $\mathbb{M}  _n $  to refer to the space of square matrices of order 
$n$. Given a matrix $A \in \mathbb{M}  _{n , m} $, we denote its components by $A _{ij} $ and we write $A=(A_{ij})$, with $i \in \left\{ 1, \dots, n\right\} $, $j \in \left\{ 1, \dots m\right\} $.   If $A$ and $B$  are two matrices with the same number of rows, we denote by $(A|| B) $ the matrix resulting from their horizontal concatenation. We write $\mathbb{I} _n $  to denote the identity matrix  of dimension $n$.
We use $\mathbb{S}_n   $ to  indicate the subspace $\mathbb{S}  _n \subset \mathbb{M}  _n $ of symmetric matrices, that is, $\mathbb{S}  _n = \left\{ A \in \mathbb{M}  _n \mid A ^\top = A\right\}$.
The symbol $||A||_{\rm Frob}$ denotes the Frobenius norm of $A \in \mathbb{M} _{m,n}$ defined  as $\|A\|_{{\rm Frob}}^2:={\rm trace} \left(A^T A\right) $~\cite{Meyer:book:matrix}.  The symbols ${\rm E}[\cdot]$ and ${\rm Cov}(\cdot,\cdot)$ denote the expectation and the covariance of random variables, respectively. Given a random variable $X$ the symbols ${\rm E}_X[\cdot]$ and ${\rm Cov}_X(\cdot,\cdot)$ denote  the conditional expectation and the conditional covariance with respect to $X$, respectively. Given a random vector $\mathbf{x} $ and a random matrix $X$, we will  use the symbols $\boldsymbol{\mu}_{\mathbf{x}} $ and $M_X$ to denote the mean of $\mathbf{x} $ and $X$, respectively. Let $Z$ be a $m$ by $n$ matrix random variable; the symbol $Z\sim {\rm MN} _{m,n}(M_Z, U_Z, V_Z)$ indicates that $Z$ is distributed according to the matrix valued  normal distribution with mean matrix $M_Z\in \Bbb M _{m,n}$ and   scale matrices   $U_Z\in \Bbb S _{m}$ and $V_Z\in \Bbb S _{n}$ (see \cite{bookMatrixDistributions2000} for additional definitions and properties).

\medskip

\noindent {\bf Glossary of symbols.} $p$ is the number of explanatory variables (dimension of $\mathbf{x}$), $q$ is the number of dependent variables (dimension of $\mathbf{y}$), $N$ is the dimension of the (random) sample size, $\lambda$ is the regularization strength.

\noindent $\boldsymbol{\mu} _{\mathbf{x}}:= {\rm E}\left[\mathbf{x}\right]$, $\boldsymbol{\mu} _{\mathbf{y}}:= {\rm E}\left[\mathbf{y}\right]$, $\Sigma _{\bf x} := {\rm Cov} ( {\bf x} , {\bf x} ) \in \Bbb S _p $, $\Sigma _{\bf y} := {\rm Cov} ( {\bf y} , {\bf y} ) \in \Bbb S _q$, $\Sigma _{{\bf x}{\bf y}} := {\rm Cov} ( {\bf x} , {\bf y} ) \in \Bbb M _{p, q}$.

\noindent$B _\lambda := \left(\Sigma _{\bf x}  + \lambda \mathbb{I}_p \right) ^{-1}  \Sigma _{\bf xy}$ is the ridge regression matrix.

\noindent $\boldsymbol{\varepsilon} _\lambda := {\bf y} - B _\lambda^\top {\bf x} $ are the ridge regression residuals.

\noindent $\boldsymbol{\mu} _{\boldsymbol{\varepsilon}_\lambda}: ={\rm E}\left[ \boldsymbol{\varepsilon} _\lambda   \right] = \boldsymbol{\mu} _{\mathbf{y}}  - B _\lambda ^\top \boldsymbol{\mu} _{\mathbf{x}} $, $\boldsymbol{\mu} _{\boldsymbol{\varepsilon}_\lambda | {\bf x} }: ={\rm E} _\mathbf{x}\left[ \boldsymbol{\varepsilon} _\lambda   \right] = {\rm E}_\mathbf{x}\left[ \mathbf{y}  \right]  - B _\lambda ^\top \mathbf{x} $.

\noindent $\Sigma _{ {\boldsymbol{\varepsilon}} _\lambda} :={\rm Cov} \left( \boldsymbol{\varepsilon} _\lambda, \boldsymbol{\varepsilon} _\lambda\right)$, $\Sigma _{ {\boldsymbol{\varepsilon}} _\lambda|{\bf x}} :={\rm Cov} _{\bf x}\left( \boldsymbol{\varepsilon} _\lambda, \boldsymbol{\varepsilon} _\lambda\right).$

\noindent $X:=\left( \mathbf{x} _1 ||\mathbf{x} _2 || \dots ||\mathbf{x} _N \right) \in \mathbb{M} _{p, N}$, $Y:=\left( \mathbf{y} _1 ||\mathbf{y} _2 || \dots ||\mathbf{y} _N \right) \in \mathbb{M} _{q, N}$, and $E _\lambda:=\left( \boldsymbol{\varepsilon}  _{\lambda  ,1}|| \dots||\boldsymbol{\varepsilon}  _{\lambda  ,N} \right) \in \mathbb{M} _{q, N}$.

\noindent $A_N:= \mathbb{I} _N - \dfrac{1}{N} \mathbf{i} _N \mathbf{i} _N ^\top $.
 
\noindent$ \widehat{B} _\lambda :=(XA_NX^\top + \lambda N \mathbb{I}_p ) ^{-1}  XA_NY^\top$ is the ridge regression matrix estimator.

\noindent $Z _\lambda := R _{\lambda} XA_N X ^\top=\mathbb{I}_p - \lambda N R _{\lambda}$.

\noindent $
R _{\lambda}:= (XA _N X ^\top + \lambda N \mathbb{I}_p ) ^{-1}.
$

\noindent For $XA _NX^\top$ invertible:

\noindent$ \widehat{B} :=\widehat{B}_0=(XA_NX^\top) ^{-1}  XA_NY^\top$ is the ordinary least squares matrix estimator.
 
\noindent $\widehat{B} _\lambda = Z _\lambda \widehat{B}$.

\noindent$Z _\lambda := R _{\lambda} XA_N X ^\top=\mathbb{I}_p - \lambda N R _{\lambda}= \left( \mathbb{I} _p + \lambda N(XA _N X ^\top \right) ^{-1} ) ^{-1}$.

\medskip

\noindent {\bf Acknowledgments:} We acknowledge partial financial support of the French ANR ``BIPHOPROC" project (ANR-14-OHRI-0002-02).

\section{Multivariate ridge regressions and the ridge estimator}
\label{Multivariate linear ridge regression and ridge estimator}
\subsection{The setup} 
\label{The covariates and the explained variables}

\medskip

Let $ {\bf x} $ and $ {\bf y} $ be respectively $\mathbb{R} ^p $ and $\mathbb{R} ^q $-valued random variables defined on a given probability space $( \Omega , \mathcal{F}, \mathbb{P} )$. All along this paper we work under the following assumption:

\medskip

\noindent {\bf (A1)}  {\it The random variables $ {\bf x},  {\bf y}$ belong to $ {L}^2 ( \Omega , \mathcal{F}, \mathbb{P} )$, that is  their first and second order moments exist and are finite, that is, ${\rm E} \left[ {\bf x} \right] < \infty$, ${\rm E} \left[ {\bf y} \right]  < \infty$,  ${\rm E} \left[ {\bf x}{\bf x}^\top  \right] < \infty$, ${\rm E} \left[ {\bf y}  {\bf y}^\top \right]  < \infty$. Additionally, the Cauchy-Schwarz inequality implies the finiteness of the covariance between  $ {\bf x}$ and $ {\bf y}$,  that is, ${\rm Cov} \left( {\bf x},{\bf y}  \right) < \infty$.}

\medskip

\noindent In the sequel we denote $\boldsymbol{\mu} _{\bf x}:={\rm E} \left[ {\bf x} \right] \in \mathbb{R} ^p $,  $\boldsymbol{\mu} _{\bf y}:={\rm E} \left[ {\bf y} \right] \in\mathbb{R} ^q $ and use the following notation for the second order central moments: $\Sigma _{\bf x} := {\rm Cov} ( {\bf x} , {\bf x} ) \in \Bbb S _p $, $\Sigma _{\bf y} := {\rm Cov} ( {\bf y} , {\bf y} ) \in \Bbb S _q$, and  $\Sigma _{{\bf x}{\bf y}} := {\rm Cov} ( {\bf x} , {\bf y} ) \in \Bbb M _{p, q}$. 

\medskip

\noindent {\bf The multivariate  ridge regression optimization problem.} Consider the ridge penalized least squares problem and define:
\begin{align}
\label{ridge optimization problem statement}
{ B}_{\lambda }&:=\mathop{\rm arg\, min}_{{ B}   \in \mathbb{M}_{p,q} } \left({\rm trace} \left({\rm E} \left[  ({ B} ^{ \top}  {\bf x}    - {\bf y}   ) ({ B} ^{ \top}  {\bf x}    -{\bf y}   )^\top \right] \right)  + \lambda \|{ B}   \|^2_{\rm Frob}\right),
\end{align}
where $ \lambda \in \mathbb{R} ^+$ is regularization strength and $||\cdot||^2_{\rm Frob}$ denotes the squared Frobenius norm of the  coefficient matrix $B$. The optimization problem \eqref{ridge optimization problem statement}  does not contain an intercept which does not imply any loss of generality since its presence can be accommodated by replacing   $B \in \mathbb{M} _{p,q}$ and  $\mathbf{x} \in \mathbb{R} ^p $ in~\eqref{ridge optimization problem statement} by $\tilde{B} \in \mathbb{M} _{p + 1,q}$ and  $\tilde{\mathbf{x}} \in \mathbb{R} ^{p+1} $, respectively,  that are constructed by vertical concatenation of  the  intercept vector ${\bf b}_0^\top \in \mathbb{R} ^q $  with $B$ and of $1$ with $\mathbf{x} \in \mathbb{R} ^p $, respectively. In this case the penalization summand in \eqref{ridge optimization problem statement} has to be replaced by $\lambda \|{ \tilde{B}}   \|^{2 ^\ast  }_{\rm Frob}$ where $|| \tilde{B}||^{2 ^\ast  }_{\rm Frob}$ would denote the  squared Frobenius norm of the $p \times  q$ lower submatrix of  $\tilde{B}$. 

In the presence of the assumption {\bf (A1)} the multivariate ridge regression problem \eqref{ridge optimization problem statement} has a unique solution $B_\lambda \in \mathbb{M} _{p,q}$  given by
\begin{equation}
\label{B lambda}
B _\lambda := \left(\Sigma _{\bf x}  + \lambda \mathbb{I}_p \right) ^{-1}  \Sigma _{\bf xy}.
\end{equation}
We will refer to $B _\lambda $ as the ridge regression matrix. We emphasize that   $B _\lambda$ always exists even if the covariance matrix $\Sigma _{\bf x}$ is singular or ill-conditioned, provided that $\lambda > 0 $.

\medskip

\noindent {\bf The regression residuals.} Given the random variables $ {\bf x},  {\bf y}$ that satisfy the hypothesis {\bf (A1)}, we can define, for each $\lambda \in \mathbb{R}^+ $, the regression residuals using the corresponding and uniquely determined ridge regression matrix $B _\lambda $ in~\eqref{B lambda} as:
\begin{equation}
\label{regression residuals}
\boldsymbol{\varepsilon} _\lambda := {\bf y} - B _\lambda^\top {\bf x}.
\end{equation}
The residuals $\boldsymbol{\varepsilon} _\lambda \in \mathbb{R} ^q $ in \eqref{regression residuals} are a $\mathbb{R}^q$-valued distributed random variable with  mean $\boldsymbol{\mu} _{\boldsymbol{\varepsilon}_\lambda } \in \mathbb{R} ^q $ and covariance matrix $\Sigma _{ {\boldsymbol{\varepsilon}} _\lambda} \in \Bbb S ^q  $, that is,
\begin{equation}
\label{distr eps}
\boldsymbol{\varepsilon} _\lambda \sim D( \boldsymbol{\mu} _{\boldsymbol{\varepsilon}_\lambda }, \Sigma _{ {\boldsymbol{\varepsilon}} _\lambda}),\ \mbox{with} \ \boldsymbol{\mu} _{\boldsymbol{\varepsilon}_\lambda } :=\boldsymbol{\mu} _{\bf y} - B _\lambda ^\top \boldsymbol{\mu} _{{\bf x}},\  \Sigma _{ {\boldsymbol{\varepsilon}} _\lambda}:=\Sigma_{\bf y} - B _\lambda ^\top \Sigma _{\bf xy} - \Sigma _{\bf xy}  B _\lambda + B _\lambda ^\top (\Sigma _{\bf x}+ \boldsymbol{\mu} _{{\bf x}} \boldsymbol{\mu} _{{\bf x}}^\top ) B _\lambda.
\end{equation}
In the following sections we will also use the conditional expectation and variance of  the residuals $\boldsymbol{\varepsilon} _\lambda \in \mathbb{R} ^q $ with respect to $\mathbf{x}$, that we denote by 
\begin{equation}
\label{mu eps lambda cond x}
\boldsymbol{\mu} _{\boldsymbol{\varepsilon}_\lambda | {\bf x} }: ={\rm E} _\mathbf{x}\left[ \boldsymbol{\varepsilon} _\lambda   \right] = {\rm E}_\mathbf{x}\left[ \mathbf{y}  \right]  - B _\lambda ^\top \mathbf{x} \quad \mbox{and} \quad \Sigma _{ {\boldsymbol{\varepsilon}} _\lambda|{\bf x}} :={\rm Cov} _{\bf x}\left( \boldsymbol{\varepsilon} _\lambda, \boldsymbol{\varepsilon} _\lambda\right),
\end{equation}
respectively. Recall that by the laws of total expectation and variance, these moments and conditional moments are related by
\begin{equation*}
\boldsymbol{\mu} _{\boldsymbol{\varepsilon}_\lambda } = {\rm E} \left[\boldsymbol{\mu} _{\boldsymbol{\varepsilon}_\lambda | {\bf x} }\right] \quad \mbox{and} \quad
\Sigma _{ {\boldsymbol{\varepsilon}} _\lambda}={\rm E}\left[\Sigma _{ {\boldsymbol{\varepsilon}} _\lambda|{\bf x}} \right]+
{\rm Cov} \left(\boldsymbol{\mu} _{\boldsymbol{\varepsilon}_\lambda | {\bf x} },\boldsymbol{\mu} _{\boldsymbol{\varepsilon}_\lambda | {\bf x} }\right).
\end{equation*}

We emphasize that in our treatment of the ridge regression problem we do not assume, as it is customary in the literature, that there is an underlying stochastically perturbed  linear functional relation between $\mathbf{x} $ and $ {\bf y} $. The aim of the error formulas that we present later on in Section~\ref{Evaluation of the ridge regression errors: training and testing} is indeed the quantitative evaluation of the inaccuracy committed when the actual functional link between  those two random variables is approximated by a  linear model obtained via the solution of the regularized regression problem~\eqref{ridge optimization problem statement}. More explicitly, we proceed by first presenting the two random variables $\mathbf{x} $ and $ {\bf y} $ that satisfy hypothesis {\bf (A1)} and by solving, for a fixed penalization strength $\lambda  $, the ridge penalized least squares problem~\eqref{ridge optimization problem statement}; as we see later on, this strategy allows for the definition of the regression residuals ${\boldsymbol{\varepsilon}} _\lambda$ and, when they satisfy certain homoscedasticity conditions, it can be used to characterize the statistical properties of the ridge estimator in fully singular situations ($\Sigma _{\bf x} $ is not invertible) and to evaluate the regression (also called training) and the generalization (also called testing) errors.

\subsection{The finite sample ridge estimator, the residuals, and the homoscedasticity hypothesis}
\label{The ridge estimator and its properties} 
We now study the finite sample properties of the ridge estimator. We start by considering two  random samples of size $N$, that is, $\left\{ \mathbf{x} _i\right\}_{i \in \left\{ 1, \dots, N \right\}} $  and $\left\{ \mathbf{y} _i\right\} _{i \in \left\{ 1, \dots, N \right\}}$.  These samples are constituted by $N$ different $ \mathbb{R} ^p $-valued (respectively, $ \mathbb{R} ^q $-valued), mutually independent, and identically distributed random variables $\mathbf{x} _i \in {L}^2( \Omega , \mathcal{F}, \mathbb{P} )$ (respectively, $\mathbf{y} _i \in {L}^2( \Omega , \mathcal{F}, \mathbb{P} )$), $i \in \left\{ 1, \dots, N \right\} $. Notice that we require that all these random variables satisfy the hypothesis {\bf(A1)}. We now horizontally  concatenate  the elements of each of these two  random samples and construct the corresponding random matrices $X \in \mathbb{M} _{p, N}$ and $Y \in \mathbb{M} _{q, N}$, respectively, that is $X:=\left( \mathbf{x} _1 ||\mathbf{x} _2 || \dots ||\mathbf{x} _N \right) $ and $Y:=\left( \mathbf{y} _1 ||\mathbf{y} _2 || \dots ||\mathbf{y} _N \right) $.

The finite sample estimator $\widehat{B}_ \lambda $ of the ridge regression matrix $B _\lambda $ is constructed by plugging in \eqref{B lambda} the natural estimators for the second order moments $\Sigma _{\bf x}$ and $\Sigma _{\bf xy}$, that is, 
\begin{align}
\label{Sigma x}
\widehat{\Sigma }_{\bf x} &:=  \dfrac{1}{N} XA_NX^\top,\\
\widehat{\Sigma }_{\bf xy} &:=  \dfrac{1}{N} XA_NY^\top, \label{Sigma xy} 
\end{align}
with \begin{equation}
\label{A N}
A_N:= \mathbb{I} _N - \dfrac{1}{N} \mathbf{i} _N \mathbf{i} _N ^\top.
\end{equation}
The expressions \eqref{Sigma x}-\eqref{Sigma xy} substituted in \eqref{B lambda} yield
\begin{equation}
\label{B lambda random sample est}
\widehat{B} _\lambda :=(XA_NX^\top + \lambda N \mathbb{I}_p ) ^{-1}  XA_NY^\top,
\end{equation}
which can be subsequently evaluated  for any given realization $\mathcal{X}$ and $\mathcal{Y}$ of the random matrices $X$ and $Y$.

Consider now the $\mathbb{R} ^q $-valued random residuals $\left\{ \boldsymbol{\varepsilon}  _{\lambda  ,i} \right\} _{i \in \left\{ 1, \dots, N \right\} } $  obtained using the random samples $\left\{ \mathbf{x}   _{i} \right\} _{i \in \left\{ 1, \dots, N \right\} } $ and $\left\{ \mathbf{y} _i  \right\} _{i \in \left\{ 1, \dots, N \right\} } $, for each $i \in \left\{ 1, \dots, N \right\} $ via the assignment $\boldsymbol{\varepsilon}  _{  \lambda , i} := \mathbf{y} _i - B _\lambda \mathbf{x}_i $. Similarly to the procedure used in the construction of the random matrices $X$ and $Y$ we  horizontally concatenate the entries $\boldsymbol{\varepsilon} _{ \lambda ,i} \in\mathbb{R} ^q $, $i \in \left\{ 1, \dots, N \right\} $, and we obtain the random matrix $E _\lambda \in \mathbb{M} _{q, N}$ given by $E _\lambda:=\left( \boldsymbol{\varepsilon}  _{\lambda  ,1}|| \dots||\boldsymbol{\varepsilon}  _{\lambda  ,N} \right) $ and that satisfies the relation $E _\lambda = Y - B_\lambda ^\top X$. We denote the unconditional first moment of $E _\lambda$ by $M_{ E _\lambda}\in \mathbb{M} _{q,N}$ and note that 
\begin{equation}
\label{mu E lambda}
M_{ E _\lambda }:={\rm E} \left[ E _\lambda \right] =  \boldsymbol{\mu} _{\boldsymbol{\varepsilon}_{\lambda} } \mathbf{i} _N^\top,
\end{equation}
where  $ \boldsymbol{\mu} _{\boldsymbol{\varepsilon}_{\lambda} }$ is given by \eqref{distr eps}. We analogously specify  the conditional expectation of $E _\lambda$ given a  random matrix $X$ as
\begin{equation}
\label{mu E lambda cond X}
M_{ E _\lambda | X}:={\rm E}_X \left[ E _\lambda \right] = (\boldsymbol{\mu} _{\boldsymbol{\varepsilon}_{\lambda, 1} | { \mathbf{x} _1 } }|| \dots|| \boldsymbol{\mu} _{\boldsymbol{\varepsilon}_{\lambda, N} | { \mathbf{x} _N } }),
\end{equation}
where each $\boldsymbol{\mu} _{\boldsymbol{\varepsilon}_{\lambda, i} | { \mathbf{x} _i } }$, $i \in \left\{ 1, \dots, N \right\} $ is given by \eqref{mu eps lambda cond x}. 

The relations that we just provided for the conditional and unconditional first order moments of the residuals random matrix $E _\lambda $ can be used to establish the expressions of their corresponding second-order counterparts. In general, one needs to use \eqref{distr eps} and \eqref{mu eps lambda cond x} in order to determine the second order moments of  the  residuals $\left\{ \boldsymbol{\varepsilon}  _{\lambda  ,i} \right\} _{i \in \left\{ 1, \dots, N \right\} } $ and this implies that, in principle,  there exist $N$ different conditional covariance matrices $  \Sigma _{ \boldsymbol{\varepsilon}  _{ \lambda, i}  | \mathbf{x} _i  }$, $i \in \left\{ 1, \dots, N \right\} $, defined for each $\boldsymbol{\varepsilon} _{ \lambda , i}$ and given some corresponding random sample element $ \mathbf{x}_i $. We consider in what follows a particular situation in which all these second order conditional moments  are constant and equal for all $i \in \left\{ 1, \dots, N \right\} $. As we discuss later on in the text, this conditional homoscedasticity property is natural and is satisfied in many standard situations. In particular, we will show that linear and nonlinear models with additive noise (examples A and B) have conditionally homoscedastic residuals unlike the case with multiplicative noise (Example C) that in general does not satisfy this property.

\begin{lemma}
\label{Lemma 2}  
In the  conditions that were just introduced consider the ridge residuals $\left\{ \boldsymbol{\varepsilon}  _{\lambda  ,i} \right\} _{i \in \left\{ 1, \dots, N \right\} }   \sim D( \boldsymbol{\mu}_{ \boldsymbol{\varepsilon}  _\lambda}, \Sigma _{  \boldsymbol{\varepsilon}  _ \lambda})$ and suppose that they are homoscedastic, that is, we assume that the conditional covariance matrices $ \Sigma _{ \boldsymbol{\varepsilon}  _{ \lambda, i}  | \mathbf{x} _i  }$, $i \in \left\{ 1, \dots, N \right\} $, in \eqref{sigma lambda x}  are constant and  equal to some matrix $\Sigma _{  \boldsymbol{\varepsilon}  | \mathbf{x} } ^ \lambda \in \Bbb S^+ _q $. 
Then the following relations hold true:
\begin{description}
\item  [{\bf (i)}] For the conditional second-order  raw moments: 
\begin{align}
\label{E Elambda tr Elambda cond X}
{\rm E}_X \left[ E _\lambda ^\top E _\lambda \right] - M_{ E _\lambda | X} ^\top M_{ E _\lambda | X} &= {\rm trace}(\Sigma _{  \boldsymbol{\varepsilon}  | \mathbf{x} } ^ \lambda) \mathbb{I} _N,\\
{\rm E} _X\left[ E _\lambda  E _\lambda^\top \right] -M_{ E _\lambda | X} M_{ E _\lambda | X}^\top&=N \Sigma _{  \boldsymbol{\varepsilon}  | \mathbf{x} } ^ \lambda, \label{E Elambda Elambda tr cond X}
\end{align}
\item  [{\bf (ii)}] For the unconditional second-order  raw moments:
\begin{align}
\label{E Elambda tr Elambda}
{\rm E} \left[ E _\lambda ^\top E _\lambda \right] - M_{E_\lambda}^\top M_{E_\lambda} &= {\rm trace}(\Sigma _{  \boldsymbol{\varepsilon}  _\lambda  } ) \mathbb{I} _N,\\
{\rm E} \left[ E _\lambda  E _\lambda^\top \right] - M_{E_\lambda} M_{E_\lambda} ^\top&=N \Sigma _{  \boldsymbol{\varepsilon}  _\lambda  } .\label{E Elambda Elambda}
\end{align}
\end{description}
In these relations, the mean $M_{ E _\lambda} $  and conditional mean $M_{ E _\lambda | X} $ matrices  are given by~\eqref{mu E lambda} and~\eqref{mu E lambda cond X}, respectively.
\end{lemma}
The proof of this lemma is not provided since it is straightforward. Lemma \ref{Lemma 2} has an important implication under an additional normality assumption for the ridge residuals that we state in the following corollary.

\begin{corollary} \label{Corollary Lemma 2} 
Suppose that the hypotheses of Lemma \ref{Lemma 2} are satisfied and that, additionally, the conditional residuals are independent and normally distributed, that is, $\boldsymbol{\varepsilon} _{\lambda, i} | \mathbf{x}_i \sim {\rm IN} ( \boldsymbol{\mu} _{ \boldsymbol{\varepsilon} _{\lambda, i}| \mathbf{x} _i }, \Sigma _{  \boldsymbol{\varepsilon}  | \mathbf{x} } ^ \lambda)$, for each $i \in \left\{ 1, \dots, N\right\} $. Then, the random matrix of residuals $E _{\lambda}$ conditional on $X$ are distributed according to a matrix normal distribution with  conditional mean $M_{ E _\lambda | X}$, and  $\Sigma _{  \boldsymbol{\varepsilon}  | \mathbf{x} } ^ \lambda$ and $\mathbb{I}_N$ as scale matrices, that is,  
\begin{equation}
\label{E lambda cond X distribution}
E _{\lambda} |X \sim {\rm MN} ( M_{ E _\lambda | X}, \Sigma _{  \boldsymbol{\varepsilon}  | \mathbf{x} } ^ \lambda, \mathbb{I}_N ).
\end{equation}
\end{corollary}

\subsection{The properties of the ridge regression estimator}
\label{The properties of the ridge regression estimator} 

The results presented in Lemma \ref{Lemma 2} and its Corollary \ref{Corollary Lemma 2} allow us to  take an important step. Indeed, we now see how, in the presence of the conditional normality and homoscedasticity hypotheses on the ridge residuals that we introduced above, we can spell out the properties of the ridge regression matrix estimator $\widehat{B} _\lambda$ introduced in~\eqref{B lambda random sample est}, conditional on the covariates sample $X$. As we already pointed out in the introduction, our results generalize the classical ones in \cite{hoerl_ridge_1970} by formulating the properties of $\widehat{B} _\lambda$ directly in terms of the solution $B _\lambda $ of the ridge regularized  regression problem and without assuming the existence of a least squares matrix coefficient $B$ or, equivalently, the regularity of the covariance matrix ${\Sigma }_{ \mathbf{x} }$ of the covariates, that in many applications is just not available.

\begin{theorem} 
\label{Theorem 1}  
Consider the random samples $\left\{ \mathbf{x} _i \right\} _{i \in \left\{ 1, \dots, N \right\} } $ and $\left\{ \mathbf{y} _i \right\} _{i \in \left\{ 1 ,\dots, N \right\} } $ of length $N$ that consist of the random variables  $\mathbf{x} _i \in \mathbb{R} ^p $, $\mathbf{y} _i \in \mathbb{R} ^q $, $i \in \left\{ 1, \dots, N \right\} $, respectively,  that satisfy the hypothesis {\bf (A1)}. Let $X\in \mathbb{R} _{p, N}$ and $Y \in \mathbb{M} _{q, N}$ be the random matrices constructed by  horizontal concatenation of all the corresponding entries of $\left\{ \mathbf{x} _i \right\} _{i \in \left\{ 1, \dots, N \right\} } $ and $\left\{ \mathbf{y} _i \right\} _{i \in \left\{ 1 ,\dots, N \right\} } $, respectively. Additionally, let $\left\{ \boldsymbol{\varepsilon}  _{  \lambda , i} \right\}  _{i \in \left\{ 1, \dots, N \right\} }$ be the associated  residuals of the ridge regression with  regularization strength $ \lambda \in \mathbb{R} ^+$ defined by $\boldsymbol{\varepsilon}  _{  \lambda , i} := \mathbf{y} _i - B _\lambda \mathbf{x}_i $ and that we assume to be conditionally homoscedastic. Let $E _\lambda $ be the random matrix obtained by horizontal concatenation of the ridge residuals conditioned on $X$ and suppose that it is normally distributed as in \eqref{E lambda cond X distribution} with conditional mean $M_{ E _\lambda | X} $ and some  constant scale  matrix $\Sigma _{  \boldsymbol{\varepsilon}  | \mathbf{x} } ^ \lambda$. Finally,  let  $\widehat{B} _\lambda $ be the ridge estimator of the ridge regression matrix $B _\lambda $ based on the random sample matrices $X$ and $Y$. Then
\begin{equation}
\label{B lambda hat minus B lambda}
(\widehat{B} _\lambda - B_\lambda )|X \sim {\rm MN} (-\lambda N R _\lambda B_\lambda + R_\lambda X A _N  M_{ E _\lambda | X} ^\top, Z _\lambda  R_\lambda ,\Sigma _{  \boldsymbol{\varepsilon}  | \mathbf{x} } ^ \lambda), 
\end{equation} 
where  the conditional mean $M_{ E _\lambda | X}$ is given in \eqref{mu E lambda cond X}, the conditional  covariance matrix $\Sigma _{  \boldsymbol{\varepsilon}  | \mathbf{x} } ^ \lambda$ is provided in point {\bf (i)} of Lemma \ref{Lemma 2} and, additionally,  
\begin{align}
\label{R lambda repeat}
R _\lambda &:=(XA _N X ^\top +\lambda N \mathbb{I}_p )^{-1},\\
Z _\lambda &:=R _\lambda X A_N X ^\top = \mathbb{I} _p - \lambda N R _\lambda \label{Z lambda B lambda}
\end{align}
with $A_N$ as in \eqref{A N}.
\end{theorem}

\subsection{Examples}

We now illustrate the different elements and developments in the paper, as well as the reach of the hypotheses that are invoked,  with three examples. Example A contains the standard linear  multivariate regression model under the Gauss-Markov hypotheses; we will show in the context of this example that the classical results in~\cite{hoerl_ridge_1970} can be obtained as a particular case of Theorem~\ref{Theorem 1}. Examples~B and C consider fully nonlinear functional relations in the generation of $ {\bf y} $ that are subjected to additive and multiplicative stochastic disturbances, respectively. A particular case of Example~B will be worked out in Section~\ref{Numerical illustration} in order to illustrate some of the consequences of the error formulas in Section~\ref{Evaluation of the ridge regression errors: training and testing}.

\medskip

\noindent $\blacktriangleright$ {\bf Example~A.  Linear  multivariate regression model}.   Consider a multivariate  linear regression model  of the form
\begin{equation}
\label{regression model}
\mathbf{y} = B^\top \mathbf{x} + \boldsymbol{\varepsilon}, 
\end{equation}
where $\mathbf{x} $ and $\mathbf{y} $ are $ \mathbb{R} ^p $ and $\mathbb{R} ^q $-valued random variables, respectively, subjected to the hypothesis {\bf (A1)}. The term $\boldsymbol{\varepsilon}$ is a $\mathbb{R} ^q $-valued  random variable  with zero mean and covariance matrix $\Sigma _{\boldsymbol{\varepsilon} }$, that is, $\boldsymbol{\varepsilon} \sim {\rm N}( {\bf 0} _q , \Sigma _{\boldsymbol{\varepsilon} })$. We place ourselves in a Gauss-Markov setup by assuming that $\boldsymbol{\varepsilon}$ and $\mathbf{x} $ are independent random variables. 
In these conditions  it is easy to show that for any $ \lambda \in \mathbb{R} ^+$, the ridge regression matrix $B _\lambda $ in \eqref{B lambda} and the linear regression coefficient matrix $B$ in \eqref{regression model} are related by 
\begin{equation}
\label{B lambda on B}
B_\lambda = ( \Sigma _{\bf x} +\lambda  \mathbb{I}_p )^{-1} \Sigma _{\bf x} B, \enspace {\rm or,} \enspace {\rm equivalently,} \enspace B_\lambda - B  = - \lambda  ( \Sigma _{\bf x} +\lambda  \mathbb{I}_p )^{-1}B,
\end{equation}
where we used in \eqref{B lambda} that $\Sigma _{\bf xy} = \Sigma _{\bf x} B$. Since the ridge residuals are determined by the equality $\mathbf{y}= B _\lambda ^\top \mathbf{x} + \boldsymbol{\varepsilon}  _\lambda $ and, at the same time, $\mathbf{x} $ and $\mathbf{y} $ are related by~\eqref{regression model}, we have that
\begin{equation*}
\boldsymbol{\varepsilon} _\lambda := \mathbf{y} - B _\lambda ^\top \mathbf{x} = (B - B _\lambda )^\top \mathbf{x}  + \boldsymbol{\varepsilon}.
\end{equation*} 
The unconditional first and the second moments of the ridge residuals $\boldsymbol{\varepsilon} _\lambda$  can be obtained in a straightforward way out of the relations  in \eqref{distr eps} as follows: 
\begin{align}
\label{mu eps lambda example}
\boldsymbol{\mu} _{\boldsymbol{\varepsilon}_\lambda }&= (B - B _\lambda ) ^\top \boldsymbol{\mu}_{\bf x},\\
\label{sigma lambda example}
\Sigma _{ \boldsymbol{\varepsilon} _\lambda} &= \Sigma _ {\boldsymbol{\varepsilon} } + (B-B_\lambda )^\top \Sigma _{\bf x} (B - B _\lambda ),
\end{align}
where we used that $\Sigma _{\bf y} = \Sigma _ { \boldsymbol{\varepsilon} } + B ^\top \Sigma _{\bf x} B$ and that $\Sigma _{\bf yx} =B^\top \Sigma _{\rm x} $.  The conditional counterparts of \eqref{mu eps lambda example}-\eqref{sigma lambda example} can be obtained with the help of  \eqref{mu eps  lambda cond x}:
\begin{align}
\boldsymbol{\mu} _{\boldsymbol{\varepsilon}_\lambda | {\bf x}} &= (B-B _\lambda ) ^\top \mathbf{x},\label{mu eps lambda cond x example}\\
\label{sigma lambda x}
\Sigma _{ \boldsymbol{\varepsilon} _\lambda|{\bf x}}&={\rm E} _{ \mathbf{x} }\left[ \boldsymbol{\varepsilon} _\lambda  \boldsymbol{\varepsilon} _{\lambda}^ \top \right] - \boldsymbol{\mu} _{\boldsymbol{\varepsilon}_\lambda | {\bf x}} \boldsymbol{\mu} _{\boldsymbol{\varepsilon}_\lambda | {\bf x}} ^\top = \Sigma _ {\boldsymbol{\varepsilon}}.
\end{align}
Notice that this expression shows that the ridge regression residuals of the linear regression model are conditionally homoscedastic. 

We now  derive  the particular expression of the ridge regression matrix estimator in the conditions implied by~\eqref{regression model}. We  first recall that the ordinary least squares estimator $\widehat{B}$ of $B$ is given by \eqref{B lambda random sample est} with  zero regularization strength,  that is $\widehat{B}=\widehat{B}_0$. More explicitly,  
\begin{equation}
\label{B lambda0 random sample est}
\widehat{B} := \widehat{B} _0 =\widehat{\Sigma} _{\bf x} ^{-1} \widehat{\Sigma} _{\bf xy}=(XA_NX^\top) ^{-1}  XA_NY^\top.
\end{equation}
It is important to underline that the ordinary least squares estimator  \eqref{B lambda0 random sample est} is available only when the sample covariance matrix $\widehat{\Sigma} _{\bf x} $ in \eqref{Sigma x} is invertible which, as we already pointed out in the introduction, is one of its main deficiencies. 

The following lemma, whose proof is in Appendix \ref{Proof of Lemma 1}, provides the relation between  the ridge regression  $\widehat{B} _\lambda$ and the ordinary least squares estimator $\widehat{B}$. We obviously restrict ourselves to the case in which the evaluation of \eqref{B lambda0 random sample est} is feasible, that is, when $\widehat{\Sigma} _{\bf x} $ is invertible. The relations that we now state are not new and are well established in the literature~\cite{hoerl_ridge_1970}. We provide them for future reference and in order to complete Example A.

\begin{lemma}
\label{Lemma 1} 
In the conditions of {Example~A} the following statements hold true:
\begin{description}
\item[{\bf (i)}] The relation between the finite sample ridge estimator $\widehat{B} _\lambda $ of $B _\lambda $ and the finite sample ordinary least squares estimator $\widehat{B}$ of  $B$ is given by
\begin{equation}
\label{B lambda ZB}
\widehat{B} _\lambda = Z _\lambda \widehat{B}
\end{equation}
where\begin{align}
\label{Z lambda}
Z _\lambda &:= R _{\lambda} XA_N X ^\top,\\
\label{R lambda}
R _{\lambda}&:= (XA _N X ^\top + \lambda N \mathbb{I}_p ) ^{-1}
\end{align}
with $A_N$ as in \eqref{A N}.
Expression \eqref{B lambda ZB} is the finite sample based analogue of the first relation in   \eqref{B lambda on B}.
\item[{\bf (ii)}] The multiplier $Z _\lambda $ in \eqref{Z lambda} can be  equivalently expressed  as  
\begin{equation}
\label{Z1}
Z _\lambda = \mathbb{I}_p - \lambda N R _{\lambda},
\end{equation} 
or 
\begin{equation}
Z _\lambda = \left( \mathbb{I} _p + \lambda N(XA _N X ^\top \right) ^{-1} ) ^{-1}, \label{Z2} 
\end{equation} 
with $R _\lambda $  in \eqref{Z1}  defined as in \eqref{R lambda}. 
\end{description} 
\end{lemma} 

\begin{remark}
\normalfont
The relations $Z _\lambda = \mathbb{I}_p - \lambda N R _{\lambda}$ in~\eqref{Z1} and $Z _\lambda = R _{\lambda} XA_N X ^\top$ in~\eqref{Z lambda} hold in a general context beyond the specific conditions of Example A and even if $XX^\top$ is singular or ill-conditioned. They are used  later on in the paper.
\end{remark}

We now show how the classical results on the ridge estimator properties in Section 4a of \cite{hoerl_ridge_1970} that assume the existence of a least squares estimate follow as a corollary of Theorem \ref{Theorem 1}. The proof of the following result is provided in Appendix \ref{Proof of Corollary of Theorem 1}. 

\begin{corollary}[Section 4a in \cite{hoerl_ridge_1970}] 
\label{Corollary Theorem 1} 
Consider the   $\mathbb{R}^q$-valued  linear regression $Y = B ^\top X + E $  obtained out of random samples of length $N$ that satisfy the relation~\eqref{regression model}. Given that, by hypothesis,  the error term satisfies $\boldsymbol{\varepsilon} \sim {\rm N}( {\bf 0} _q , \Sigma _{\boldsymbol{\varepsilon} })$, it is easy to see that the residual random matrix $E\in \mathbb{M} _{q, N}$ obtained by horizontal concatenation, is  matrix normal distributed as $E \sim {\rm MN} \left( \mathbb{O}  _q , \Sigma _{\boldsymbol{\varepsilon}}, \mathbb{I} _N  \right) $. The following statements hold:
\begin{description}
\item[{\bf (i)}] The ridge regression matrix estimator is conditionally distributed as
\begin{equation}
\label{B lambda minus B our notation}
(\widehat{B} _\lambda - B)|X \sim {\rm MN} ( - \lambda N R _\lambda B, Z_\lambda R_\lambda, \Sigma _{  \boldsymbol{\varepsilon} }), 
\end{equation}
where $R _\lambda $ and $Z _\lambda $ are defined in \eqref{R lambda repeat} and in \eqref{Z lambda B lambda}, respectively.
\item[{\bf (ii)}] If the matrix $\widehat{\Sigma}_{\mathbf{x} }$ defined in \eqref{Sigma x} is invertible, then \eqref{B lambda minus B our notation} can be rewritten as
\begin{equation}
\label{B lambda minus B}
(\widehat{B} _\lambda - B)|X \sim {\rm MN} ( - \lambda N R _\lambda B, Z_\lambda (X A _N   X ^\top) ^{-1} Z_\lambda , \Sigma _{ \boldsymbol {\varepsilon} }),
\end{equation}
with $Z _\lambda $  defined as in 
\eqref{Z lambda B lambda} or, equivalently, as in \eqref{Z2}. $\blacktriangleleft $
\end{description}
\end{corollary}

\medskip

\noindent $\blacktriangleright$ {\bf Example~B.  Nonlinear one-dimensional model with additive stochastic disturbance term}.   Consider the following data generating process for the variable ${y} $
\begin{equation}
\label{exampleB model}
{y} = f(\xi ) + {\varepsilon}, 
\end{equation}
where  $f : \mathbb{R} \rightarrow  \mathbb{R} $ is a continuous function, ${\varepsilon} \sim D_ \varepsilon ( 0 , \sigma_{ \varepsilon }^2 )$ represents a disturbance term and $\xi $ is distributed with  zero mean,  variance $ \sigma ^2 _\xi$, and probability density function $g _{ \xi }$. The random variables $\varepsilon $ and $\xi $ are assumed to be independent and subjected to {\bf (A1)}. 

We now approximate the model \eqref{exampleB model} using a univariate ridge  regression  in which the $p$ explanatory variables are the first $p$ elements $\left\{ \phi _i\right\}_{i \in \left\{ 1, \dots, p \right\}}$ of a given ordered generating set   of the function space to which the function $f$ in~\eqref{exampleB model} belongs, all of them evaluated at $\xi $. More explicitly, consider an approximate model of the following form:
\begin{equation}
\label{approximated model example B}
y = {\bf B} _\lambda ^\top \mathbf{x}  + \varepsilon _\lambda, \enspace \varepsilon_\lambda  \sim D_{ \varepsilon _\lambda }( \mu _{ \varepsilon_\lambda }, \sigma _{ \varepsilon_\lambda }^2),
\end{equation}
 where  the vector of regressors $\mathbf{x} \in \mathbb{R} ^p $ is constructed as
\begin{equation}
\label{x example B}
\mathbf{x} := \left( \widetilde{\xi_1}, \dots, \widetilde{ \xi _p}  \right) ^\top, \enspace {\rm with }\enspace x _i = \widetilde{ \xi _i} := \dfrac{ \xi _i - {\rm E} \left[ \xi _i\right] }{ \sigma (\xi _i)}, \enspace \xi _i = \phi _i  ( \xi ), \enspace  i \in \left\{ 1, \dots,  p\right\}.
\end{equation}
We emphasize that for each $i \in \left\{ 1, \dots,  p\right\} $, the  explanatory variable $x _i $ is constructed out of the standardized version of the evaluation $\xi _i = \phi _i  ( \xi ) $ of the generating function associated $\phi _i$ using the standard deviation:
\begin{equation}
\label{std}
\sigma (\xi _i):= ( {\rm E} \left[ \xi _i^ {2}\right] -  {\rm E} \left[ \xi _ {i}\right]^2)^{1/2}, \enspace {\rm for} \enspace {\rm each} \enspace i\in \left\{ 1, \dots, p \right\}.
\end{equation}
We now provide all the elements associated to the approximate model~\eqref{approximated model example B}. First,~\eqref{B lambda} determines, for any regularization strength $\lambda \in \mathbb{R} ^+$, the ridge regression matrix $B _\lambda $ once the central moments $\Sigma _{\bf x} \in \Bbb S _p $ and $ {\mathbf{\Sigma}} _{{\bf x}y} \in \mathbb{R}^p $ have been computed. It is easy to see that
\begin{align}
\label{sigma x example B}
(\Sigma _{\bf x})_{i,j}=&{\rm E}\left[\widetilde{ \xi _i}\widetilde{ \xi _j} \right] = \dfrac{1}{\sigma (\xi _i)\sigma (\xi _j)}\int_{-\infty}^{\infty} \left(  \phi _i (z) - {\rm E} \left[ \xi _i\right]\right) \left( \phi _j (z) - {\rm E} \left[ \xi _j\right]\right) g_{\xi }(z) dz, \enspace i, j\in \left\{ 1, \dots, p \right\}, \\
(\mathbf{\Sigma} _{{\bf x}y})_{i}=&{\rm E}\left[\widetilde{ \xi _i}f(\xi) \right] = \dfrac{1}{\sigma (\xi _i)}\int_{-\infty}^{\infty} \left(  \phi _i (z) - {\rm E} \left[ \xi _i\right]\right) f(z) g_{\xi }(z) dz, \enspace i\in \left\{ 1, \dots, p \right\}, \label{sigma xy example B} 
\end{align}
with  $\sigma ( \xi_i)$ as in \eqref{std} and where $g_{ \xi }$ is the probability density function of $ \xi $.
We now  study the defining features of the ridge regression residuals $ \varepsilon _\lambda $ and their moments. First, by  \eqref{approximated model example B} and \eqref{exampleB model}, the residuals $ \varepsilon _\lambda$ are given by
\begin{equation*}
\varepsilon _\lambda  = f( \xi ) - {\bf B}_\lambda ^\top \mathbf{x} + \varepsilon.
\end{equation*}
This expression can be used to explicitly compute the unconditional and conditional first and second moments of  $ \varepsilon _\lambda $; indeed, by~\eqref{distr eps}:
\begin{align}
\label{mu eps lambda example B}
\mu _{\varepsilon _\lambda } &= {\rm E} \left[ f( \xi )\right] = \int _{-\infty}^{\infty} f(z) g_{ \xi } (z) dz,
\end{align}
where  we used that ${\rm E}\left[ \mathbf{x} \right] =0$ by construction in \eqref{x example B}. Analogously,  \eqref{distr eps} yields
\begin{align}
\sigma ^2 _{\varepsilon_\lambda } &= \sigma ^2 _{y} - {\bf B}_\lambda ^\top \mathbf{ \Sigma }_{\mathbf{x} y} - \mathbf{ \Sigma }_{\mathbf{x} y} ^\top {\bf B}_\lambda + {\bf B}_\lambda ^\top \Sigma _{ \mathbf{x} } {\bf B}_\lambda - \mu _{\varepsilon _\lambda }^2, \label{sigma eps lambda example B} 
\end{align}
where 
\begin{equation*}
\sigma ^2 _{y} = {\rm E} \left[ f (\xi) ^2  \right] + \sigma _{ \varepsilon } ^2
\end{equation*}
with
\begin{equation}
\label{Efx2}
{\rm E} \left[ f (\xi) ^2  \right] = \int_{-\infty}^{\infty} f(z)^2 g_{ \xi }(z) dz.
\end{equation}
Regarding the conditional moments, by \eqref{mu eps lambda cond x}  we have
\begin{align}
\mu _{\varepsilon _\lambda | \mathbf{x} } &= f ( \xi ) - {\bf B} _\lambda ^\top\mathbf{x} ,\nonumber\\
\sigma ^2 _{\varepsilon_\lambda  | \mathbf{x} } &= {\rm E}_ \mathbf{x}\left[ \varepsilon _\lambda ^2 \right] - \mu _{\varepsilon _\lambda | \mathbf{x} }^2\nonumber\\ 
&= f( \xi ) ^2 + \sigma _\varepsilon ^2 +  {\bf B}_\lambda ^\top \mathbf{x} \mathbf{x} ^\top  {\bf B}_\lambda - 2 f ( \xi )  {\bf B}_\lambda ^\top \mathbf{x} - f( \xi ) ^2 -  {\bf B}_\lambda ^\top \mathbf{x} \mathbf{x} ^\top  {\bf B}_\lambda  + 2 f ( \xi )  {\bf B}_\lambda ^\top \mathbf{x} = \sigma _\varepsilon ^2. \label{sigma eps lambda cond x example B} 
\end{align}
Notice that the expression \eqref{sigma eps lambda cond x example B} proves the conditional homoscedasticity of the ridge residuals in this setup. This observation has important implications and we hence frame it in the following proposition.

\begin{proposition}
\label{Exampe B homoscdasticity}  
Let $X$ and $Y$ be random sample matrices generated by the regression model~\eqref{approximated model example B} in Example B that assumes that  $ y$ is an additive stochastic perturbation of some continuous function of some random variable $\xi $. If, additionally, the stochastic perturbation and the random variable $\xi $ are independent, then the ridge regression residuals of~\eqref{approximated model example B}  are always conditionally homoscedastic. $\blacktriangleleft$
\end{proposition}

\medskip

\noindent$\blacktriangleright$ {\bf Example C.  Non-linear one-dimensional model with mutiplicative stochastic disturbance term}.   Consider the following data generating process for the variable ${y} $
\begin{equation}
\label{exampleC model}
{y} = f(\xi )  {\varepsilon}, 
\end{equation}
where  ${\varepsilon} \sim D_ \varepsilon ( \mu _\varepsilon  , \sigma_{ \varepsilon }^2 )$ is a disturbance term and $\xi $ is distributed with  zero mean,  variance $ \sigma ^2 _\xi$, and probability density function $g _{ \xi }$. As in Examples~A and  B, the random variables $\varepsilon $ and $\xi $ are assumed to be independent. In the conditions of this example we choose to approximate the model \eqref{exampleC model} by the same linear  regression model \eqref{approximated model example B} with the covariates in \eqref{x example B} that are used in Example B. Notice that in this case, the relations \eqref{std}-\eqref{sigma xy example B} in Example B hold true. As for the ridge regression residuals $ \varepsilon _\lambda $, the choice \eqref{exampleC model} for the data generating process implies that
\begin{equation*}
\varepsilon _\lambda  = f( \xi )\varepsilon - {\bf B}_\lambda ^\top \mathbf{x}.
\end{equation*}
It is easy to verify, using  the independence of $\xi $ and $\varepsilon $, that  the  unconditional first moment of the ridge regression residuals is 
\begin{equation*}
\mu _{\varepsilon _\lambda } = {\rm E} \left[ f( \xi ) \right] \mu _\varepsilon =\mu _\varepsilon  \int _{-\infty}^{\infty} f(z) g_{ \xi } (z) dz,
\end{equation*}
As to the unconditional second moment, the relation \eqref{sigma eps lambda example B} holds true with $\sigma ^2 _{y} $ determined by
\begin{equation}
\label{sigma y Example C}
\sigma ^2 _{y} = {\rm E} \left[ f (\xi) ^2  \varepsilon ^2 \right] - \mu _{\varepsilon _\lambda }^2 ={\rm E} \left[ f (\xi) ^2 \right] (\sigma ^2 _{ \varepsilon } + \mu _\varepsilon^2 )- \mu _{\varepsilon _\lambda }^2 
\end{equation}
and with ${\rm E} \left[ f (\xi) ^2 \right]$ as in \eqref{Efx2}.
The conditional first and second moments are determined by
\begin{align}
\mu _{ \varepsilon _\lambda | \mathbf{x} } = &f( \xi ) \mu  _\varepsilon  - {\bf B}_\lambda ^\top \mathbf{x},\nonumber \\
\label{sigma eps lambda cond x example C}
\sigma ^2  _{ \varepsilon _\lambda | \mathbf{x} }=  &{\rm E}_ \mathbf{x}\left[ \varepsilon _\lambda ^2 \right] - \mu _{\varepsilon _\lambda | \mathbf{x} }^2= f( \xi ) ^2  (\sigma ^2 _{ \varepsilon } + \mu _\varepsilon^2 ) - 2 \mu _\varepsilon f (\xi) {\bf B}_\lambda ^\top \mathbf{x}  +  {\bf B}_\lambda ^\top \mathbf{x} \mathbf{x} ^\top  {\bf B}_\lambda\nonumber\\
&  - f( \xi ) ^2\mu _\varepsilon^2 +2 \mu _\varepsilon f (\xi) {\bf B}_\lambda ^\top \mathbf{x} -  {\bf B}_\lambda ^\top \mathbf{x} \mathbf{x} ^\top  {\bf B}_\lambda= f( \xi ) ^2  \sigma _\varepsilon ^2,
\end{align}
respectively.
Relation  \eqref{sigma eps lambda cond x example C} shows that, in this case, the ridge residuals are not conditionally homoscedastic. 
$\blacktriangleleft$

\section{Evaluation of the ridge regression and generalization errors}
\label{Evaluation of the ridge regression errors: training and testing}

In this section we use the properties of the ridge estimator that we spelled out in Theorem~\ref{Theorem 1} in order to write down explicit expressions that allow for the evaluation of the regression (also called training) and the generalization (also called testing) errors committed by a regularized regression model whose coefficient $\widehat{B} _\lambda $ has been estimated using a finite sample of a given size. The regression or training error is the one committed by the regression model when evaluated with the sample that has been used to obtain $\widehat{B} _\lambda $; for the generalization or testing error, we keep $\widehat{B} _\lambda $ and we evaluate the error committed by the corresponding regression model using another sample that may have different size or even different statistical properties. As we will see, in both cases our error formulas incorporate the error committed at the time of parameter estimation.

These two errors are of much importance in specific applications involving a finite sample framework  since their relative values give indications about the pertinence of various modeling choices. Indeed, a typical behavior in a regression model is that an increase in the number of covariates makes smaller the training error. The testing error follows the same trend up to a point in which it starts increasing; that turning point in the testing error has to do with the appearance of overfitting in the training, that is, the overabundance of regressors is fitting not only the underlying deterministic model but also the noise that is perturbing it. As we will see in the example that we work out in Section~\ref{Numerical illustration}, the formulas presented in this section can be used to avoid this phenomenon at the time of deciding on the model architecture.
 
All along this section we use the same hypotheses and notation as in Theorem \ref{Theorem 1} unless explicitly stated otherwise.

\subsection{The characteristic errors of the regression model}
\label{Characteristic errors of the regression model} 
We define the characteristic error as the one committed by the regression model  without taking into account estimation errors, that is, the characteristic error is evaluated assuming that the model parameters are known. We separately address two instances of the  characteristic error. First, we consider the standard unconditional characteristic error corresponding to the ridge model with a given regularization strength $\lambda \in \mathbb{R} ^+$. This error  corresponds to the expected regression error for arbitrary realizations of the explanatory  and dependent variables $\mathbf{x}   $ and $\mathbf{y} $, respectively; it is sometimes referred to in the literature as the irreducible error (see for example \cite{Elements:learning:book}). Second, we assume that the dependent variables $\mathbf{x}   $ are fixed and we compute the characteristic error conditional on those values.

\medskip

\noindent {\bf Characteristic (irreducible) error.} It is given by:
\begin{equation}
\label{MSE}
{\rm MSE}_{\rm char} ^\lambda :={\rm E} \left[ (\mathbf{y}  - B _\lambda \mathbf{x}  )^\top (\mathbf{y}  - B _\lambda \mathbf{x}  )\right] = {\rm E} \left[ \boldsymbol{\varepsilon}  _{ \lambda }^\top \boldsymbol{\varepsilon} _{ \lambda }\right] = {\rm trace} ( \Sigma _{ {\boldsymbol{\varepsilon}} _\lambda} + \boldsymbol{\mu} _{\boldsymbol{\varepsilon}_\lambda }\boldsymbol{\mu} _{\boldsymbol{\varepsilon}_\lambda }^\top),
\end{equation}
where $\boldsymbol{\mu} _{\boldsymbol{\varepsilon}_\lambda }$ and $ \Sigma _{ {\boldsymbol{\varepsilon}} _\lambda}$ are given in \eqref{distr eps}. 
\medskip

\noindent {\bf Conditional characteristic error.} It is the characteristic error associated to the ridge regression model conditional on the covariates value $\mathbf{x} $: 
\begin{equation}
\label{MSE cond}
{\rm MSE}_{{\rm char}| \mathbf{x} } ^\lambda :={\rm E}_{\mathbf{x}} \left[ (\mathbf{y}  - B _\lambda \mathbf{x}  )^\top (\mathbf{y}  - B _\lambda \mathbf{x} )  \right] =  {\rm E}_{ \mathbf{x} } \left[ \boldsymbol{\varepsilon}_\lambda ^\top \boldsymbol{\varepsilon}_\lambda\right]  = {\rm trace} ( \Sigma _{\boldsymbol{\varepsilon} _\lambda |\mathbf{x} }+\boldsymbol{\mu} _{\boldsymbol{\varepsilon}_\lambda | {\bf x} }\boldsymbol{\mu} _{\boldsymbol{\varepsilon}_\lambda | {\bf x} }^\top),
\end{equation}
with $\boldsymbol{\mu} _{\boldsymbol{\varepsilon}_\lambda | {\bf x} }$ and $ \Sigma _{\boldsymbol{\varepsilon} _\lambda |\mathbf{x} }$ as in \eqref{mu eps lambda cond x}.
 
\medskip

\noindent We now evaluate these errors for the examples that we consider along the paper.

\medskip

\noindent$\blacktriangleright$ {\bf Example~A (continued)}. 
\normalfont  
The expressions for the characteristic and the conditional characteristic errors in \eqref{MSE} and in \eqref{MSE cond}  for the standard regression model in \eqref{regression model} can be made explicit by using the particular form of the covariance matrix $\Sigma _{ {\boldsymbol{\varepsilon}} _\lambda} $ provided in \eqref{sigma lambda example} and of the corresponding conditional covariance $\Sigma _{\boldsymbol{\varepsilon} _\lambda |\mathbf{x} }$ given in \eqref{sigma lambda x}. Indeed:
\begin{align}
\label{MSE char}
{\rm MSE}_{\rm char} ^\lambda &={\rm trace}( \Sigma _{\boldsymbol{\varepsilon}} + (B - B _\lambda ) ^\top(\Sigma _ {\mathbf{x} } + \boldsymbol{\mu}_{\bf x}\boldsymbol{\mu}_{\bf x}^\top)(B - B _\lambda ))= 
{\rm trace}( \Sigma _{\boldsymbol{\varepsilon}} + (B - B _\lambda ) ^\top{\rm E}\left[\mathbf{x}  \mathbf{x}^\top\right](B - B _\lambda )),
\end{align}
and for the conditional characteristic error:
\begin{align}
{\rm MSE}_{{\rm char}| \mathbf{x} } ^\lambda &={\rm trace}( \Sigma _{\boldsymbol{\varepsilon}} + (B - B _\lambda ) ^\top\mathbf{x}  \mathbf{x}^\top(B - B _\lambda )).\label{MSE char x} 
\end{align}
From these expressions it is obvious that the relative magnitudes of these characteristic and conditional characteristic errors depend on the specific value of the random variable $\mathbf{x} \in \mathbb{R} ^p $ used in the conditioning. $\blacktriangleleft$ 

\medskip

\noindent$\blacktriangleright$ {\bf Example~B (continued)}. In this particular instance  the characteristic error is:
\begin{align}
\label{MSE char example B}
{\rm MSE}_{\rm char} ^\lambda &= \sigma _{ \varepsilon } ^2  + {\rm E} \left[ f (\xi) ^2  \right]   + {\bf B}_\lambda ^\top \Sigma _{ \mathbf{x} } {\bf B}_\lambda -2 {\bf B}_\lambda ^\top \mathbf{ \Sigma }_{\mathbf{x} y},
\end{align}
where ${\rm E} \left[ f (\xi) ^2  \right]$, $\mathbf{ \Sigma }_{\mathbf{x} y}$, and $\mathbf{ \Sigma }_{\mathbf{x} }$ are given in \eqref{Efx2}, \eqref{sigma xy example B}, and \eqref{sigma x example B}, respectively. 
Additionally, the expression of the conditional characteristic error is 
\begin{align}
{\rm MSE}_{{\rm char}| \mathbf{x} } ^\lambda &= \sigma _\varepsilon ^2 + f (\xi) ^2 + {\bf B}_\lambda^\top \mathbf{x} \mathbf{x} ^\top {\bf B}_\lambda - 2 f ( \xi )  {\bf B}_\lambda ^\top \mathbf{x}.\label{MSE char x example B} 
\end{align}
It can be easily seen from \eqref{MSE char example B} and \eqref{MSE char x example B} that as in Example A, any ordering  between these two  errors is possible depending on the value of the random variable $\mathbf{x} \in \mathbb{R} ^p $.
$\blacktriangleleft$

\medskip

\noindent$\blacktriangleright$ {\bf Example~C
 (continued)}. The characteristic error is:
\begin{align}
\label{MSE char example C}
{\rm MSE}_{\rm char} ^\lambda &=  \sigma ^2 _{ \varepsilon } {\rm E} \left[ f (\xi) ^2 \right]   + {\bf B}_\lambda ^\top \Sigma _{ \mathbf{x} } {\bf B}_\lambda - 2 {\bf B}_\lambda ^\top \mathbf{ \Sigma }_{\mathbf{x} y} ,
\end{align}
where ${\rm E} \left[ f (\xi) ^2  \right]$, $\mathbf{ \Sigma }_{\mathbf{x} y}$, $\mathbf{ \Sigma }_{\mathbf{x} }$, and $\mu _{\varepsilon _\lambda }$ are given in \eqref{Efx2}, \eqref{sigma xy example B}, \eqref{sigma x example B}, and \eqref{mu eps lambda example B}, respectively. The expression of the conditional characteristic error is provided by:
\begin{align}
{\rm MSE}_{{\rm char}| \mathbf{x} } ^\lambda &= \sigma _\varepsilon ^2 f( \xi ) ^2 + f( \xi ) ^2\mu _\varepsilon^2 -2 \mu _\varepsilon f (\xi) {\bf B}_\lambda ^\top \mathbf{x} +  {\bf B}_\lambda ^\top \mathbf{x} \mathbf{x} ^\top  {\bf B}_\lambda .\label{MSE char x example C} 
\end{align}
As in Examples A and B, any ordering  between these two  errors is possible depending on the value of the random variable $\mathbf{x} \in \mathbb{R} ^p $.
$\blacktriangleleft$

\subsection{The conditional total training error}
\label{Conditional total training error} 
In this section we provide an explicit expression for the total error committed when using  a multivariate ridge regularized regression model with a coefficient matrix that has been estimated using a finite sample. In that situation there are two sources of error: first, the characteristic error associated to the regression model and second, the estimation error on the ridge regression matrix that appears due to the finiteness of the estimation sample. We first focus on the total regression or training error in which the error is evaluated  using  the  same sample that has been earlier exploited to obtain an estimate of the the ridge regression matrix. The expression for the error that we provide is conditional on the covariates sample used at the time of estimation.

Consider the random matrices $X \in \mathbb{M} _{p, N}$, $Y\in \mathbb{M} _{q, N}$  constructed according to the prescriptions in Section~\ref{The ridge estimator and its properties} and let $\widehat{B} _\lambda \in \mathbb{M} _{p,q}$ be the estimator of $B_\lambda $ based on $X$ and $Y$ that we introduced in~\eqref{B lambda random sample est}. The conditional total training error is defined as 
\begin{equation}
\label{MSE training}
{\rm MSE}_{{\rm training}|X}^\lambda := \dfrac{1}{N} {\rm trace} \left( {\rm E} _X\left[ (Y - \widehat{B} _\lambda ^\top X) ^\top (Y - \widehat{B} _\lambda ^\top X)\right] \right) .
\end{equation}
The following theorem, whose  proof is given in Appendix~\ref{Proof of Theorem 2}, provides an explicit expression for the conditional total training error.
\begin{theorem}
\label{Theorem 2}
Suppose that we are in the hypotheses of Theorem \ref{Theorem 1}, that is, we consider a ridge regression between square summable explanatory and dependent variables that exhibits conditionally normal and homoscedastic residuals. Then, the total training error conditional on a random sample $X$ of length $N$  of the covariates  is:
 \begin{align}
\label{MSE longer}
{\rm MSE}^\lambda&_{{\rm training}|X} = \nonumber\\
=&{\rm trace} (\Sigma _{  \boldsymbol{\varepsilon}  | \mathbf{x} } ^ \lambda ) + \dfrac{1}{N} \Big\{{\rm trace}(\Sigma _{  \boldsymbol{\varepsilon}  | \mathbf{x} } ^ \lambda) {\rm trace}\left( Z _\lambda  (R_\lambda X X ^\top - 2 \mathbb{I}_p )\right) + {\rm trace} (  M _{(\widehat{B} _\lambda -{B} _\lambda) }M _{(\widehat{B} _\lambda -{B} _\lambda) }^\top X X ^\top ) \nonumber\\
& -2 \  {\rm trace}\Big( (X ^\top R _\lambda X A_N  - \dfrac{1}{2} \mathbb{I}_N )M _{E_\lambda |X} ^\top M _{E_\lambda |X}+  (X ^\top R _\lambda X A _N  -\mathbb{I}_N ) X^\top B _\lambda \boldsymbol{\mu} _{E_\lambda |X}  \Big) \Big\} ,
\end{align}
or, equivalently,
 \begin{align}
\label{MSE shorter}
{\rm MSE}^\lambda&_{{\rm training}|X} = \nonumber\\=&{\rm trace} (\Sigma _{  \boldsymbol{\varepsilon}  | \mathbf{x} } ^ \lambda ) + \dfrac{1}{N} \Big\{{\rm trace}(\Sigma _{  \boldsymbol{\varepsilon}  | \mathbf{x} } ^ \lambda) {\rm trace}\left( Z _\lambda  (R_\lambda X X ^\top - 2 \mathbb{I}_p )\right)  + {\rm trace} \left(  M _{(\widehat{B} _\lambda -{B} _\lambda) }M _{(\widehat{B} _\lambda -{B} _\lambda) }^\top X X ^\top \right)\nonumber\\
&-2 \ {\rm trace}((X ^\top R _\lambda X A_N  - \dfrac{1}{2} \mathbb{I}_N ) M_{E_\lambda |X} ^\top M _{E_\lambda |X})\Big\} + {2}\lambda{\rm trace}(   X ^\top R _\lambda  B _\lambda M_{E_\lambda |X} ),
\end{align}
 where 
 \begin{equation}
\label{mu M}
M _{(\widehat{B} _\lambda -{B} _\lambda) }:=-\lambda N R_\lambda B_\lambda + R_\lambda X A_N  M _{E_\lambda |X} ^\top
\end{equation}
 is the bias of the estimator $\widehat{B} _\lambda $ of $B_\lambda $ in \eqref{B lambda hat minus B lambda} and where  \begin{align}
\label{Z lambda theorem}
Z _\lambda &= R _\lambda X A _N X^\top= \mathbb{I} _p - \lambda N R _\lambda,\\
R _\lambda&=\left(X  A_{N} X  ^\top + \lambda N  \mathbb{I}_p \right) ^{-1}.\label{R train}
\end{align}
\end{theorem}

\medskip

\noindent$\blacktriangleright$ {\bf Example~A (continued).} In the conditions of the linear regression model~\eqref{regression model} of Example~A,  the conditional total training error  in \eqref{MSE training} can be made more explicit by using the expressions for  the conditional covariance matrix $\Sigma _{  \boldsymbol{\varepsilon}  | \mathbf{x} } ^ \lambda $ in~\eqref{sigma lambda x} that shows that the homoscedasticity hypothesis is satisfied in this setup. Moreover, as we saw in Corollary~\ref{Corollary Theorem 1},  the properties of the estimator $\widehat{B} _\lambda $ can be formulated in this case in terms of the matrix $B$. The proof of the following result is provided in Appendix \ref{Proof of Corollary error Example A}.
\begin{corollary} 
\label{Corollary error Example A} 
In the conditions of Example~A,  the conditional total training error~\eqref{MSE training} is given by the expression:
\begin{align}
\label{MSE training example A}
{\rm MSE}_{{\rm training}|X}^\lambda = &{\rm trace} ( \Sigma _ { \boldsymbol{\varepsilon} } ) + \dfrac{1}{N} {\rm trace}(\Sigma _ { \boldsymbol{\varepsilon} }) {\rm trace}\left( Z _\lambda  (R_\lambda X X ^\top - 2 \mathbb{I}_p )\right) +  \lambda ^2 N {\rm trace}( R_\lambda B B ^\top   R_\lambda X X ^\top),
\end{align}
where $Z _\lambda$ is defined as in Corollary~\ref{Corollary Theorem 1}.
\end{corollary}
Expression~\eqref{MSE training example A} has been formulated for the first time in Proposition 3 of \cite{RC3} in a machine learning context.$\blacktriangleleft$
 
\subsection{The conditional total testing or generalization error}
\label{Conditional total testing (generalization) error} 

As we already pointed out in the introduction to this section, the minimization of the training error in the construction of a model does not suffice to guarantee its good out-of-sample performance due to the possible appearance of overfitting phenomena. This makes necessary the evaluation of the so called testing or generalization error that measures the performance of a given model that has been trained on a finite sample that is independent and may even have different statistical properties from the one that is used for the testing. In specific machine learning and forecasting applications it is mainly the testing error that needs to be used in order to decide on questions related to model architecture.

We proceed in the following way in order to carry this out: we assume that the training and testing are carried out on two different sets of  samples; those labeled with a one (respectively, two) are the training (respectively, testing) samples. More specifically, consider two pairs of random samples of sizes $N _1 $ and $N _2 $. The first pair $ \left\{\mathbf{x}  ^{(1)} _i \right\} $, $\left\{ \mathbf{y} _i ^{(1)} \right\} $, with elements $\mathbf{x}  ^{(1)} _i \in \mathbb{R} ^p $, $\mathbf{y}  ^{(1)} _i \in \mathbb{R} ^q $, $i \in \left\{ 1, \dots, N _1 \right\} $, will be used for training and the second one, denoted by $ \left\{\mathbf{x}  ^{(2)} _j \right\} $, $\left\{ \mathbf{y} _j ^{(2)} \right\} $ with $\mathbf{x}  ^{(2)} _j \in \mathbb{R} ^p $, $\mathbf{y}  ^{(2)} _j \in \mathbb{R} ^q $, $j\in \left\{ 1, \dots, N _2 \right\} $, will be used for testing. Suppose that all the elements in these random samples satisfy hypothesis  {\bf (A1)} and hence have finite first and second order moments. 
Additionally, assume that ridge residuals $ \left\{\boldsymbol{\varepsilon}  ^{(1)} _{ \lambda , i} \right\} _{i \in \left\{ 1, \dots, N _1 \right\} }$ associated to the training pair are   conditionally normal and homoscedastic in the sense of Lemma \ref{Lemma 2}, that is, their corresponding conditional covariance matrices are constant and equal to some matrix $\Sigma _{  \boldsymbol{\varepsilon}   | {\mathbf{x}  } } ^ {\lambda, (1)}\in \Bbb S _q $. This hypothesis can be stated as  $\boldsymbol{\varepsilon}  ^{(1)} _{ \lambda , i} | \mathbf{x} ^{(1)} _i  \sim {\rm N} (\boldsymbol{\mu} _{\boldsymbol{\varepsilon}^{(1)} _{\lambda , i}| \mathbf{x} ^{(1)} _i} , \Sigma _{  \boldsymbol{\varepsilon}   | {\mathbf{x}  } } ^ {\lambda, (1)})$ for each $i \in \left\{ 1, \dots, N _1 \right\} $.

Consider now the random matrices $X _1 \in \mathbb{M} _{p, N _1 }$, $Y _1  \in \mathbb{M} _{q, N _1 }$, $X _2  \in \mathbb{M} _{p, N_2 }$, $Y_2  \in \mathbb{M} _{q, N_2 }$ obtained by horizontal concatenation of the elements in the random samples $ \left\{\mathbf{x}  ^{(1)} _i \right\} _{i \in \left\{ 1, \dots, N _1 \right\}  }$, $\left\{ \mathbf{y} _i ^{(1)} \right\} _{i \in \left\{ 1, \dots, N _1 \right\}  }$,  $ \left\{\mathbf{x}  ^{(2)} _i \right\} _{i \in \left\{ 1, \dots, N _2 \right\}  }$, and $\left\{ \mathbf{y} _i ^{(2)} \right\} _{i \in \left\{ 1, \dots, N _2 \right\}  }$ respectively. We call  $( X _1 , Y _1 )$ and  $( X _2 , Y _2 )$ the training and the testing samples, respectively. 

Let now $\widehat {B} _\lambda  $ be the ridge regression matrix estimator   of ${B} _\lambda  $ constructed using a training sample $( X _1 , Y _1 )$. In these conditions we define the conditional total testing ridge error as:
\begin{equation}
\label{MSE testing}
{\rm MSE} ^\lambda _{{\rm testing}|X_1 } = \dfrac{1}{N _2 } {\rm trace} \left( {\rm E}_{X_1} \left[\left(Y _2 - \widehat{B} _\lambda ^\top X _2 \right) ^\top \left(Y _2 - \widehat{B} _\lambda ^\top X _2 \right)   \right] \right) . 
\end{equation}
This expression measures  the  mean square error committed when training is carried out with a fixed explanatory variables training sample $X _1$  of length $N _1 $ and testing is implemented with any other testing sample  $\left(X _2 , Y _2 \right) $ of  length $N _2 $. On other words, we fix $X _1$, and we average the square errors committed when varying $Y _1 $ and the testing samples $\left(X _2 , Y _2 \right) $.  The proof of the following result is provided in Appendix~\ref{Proof of Theorem testing error}.

\begin{theorem}
\label{Theorem testing error}  
Consider two pairs of random samples $ \left\{\mathbf{x}  ^{(1)} _i \right\} $, $\left\{ \mathbf{y} _i ^{(1)} \right\} $ and $ \left\{\mathbf{x}  ^{(2)} _j \right\} $, $\left\{ \mathbf{y} _j ^{(2)} \right\} $, with  $\mathbf{x}  ^{(1)} _i, \mathbf{x}  ^{(2)} _i \in \mathbb{R} ^p $, $\mathbf{y}  ^{(1)} _i, \mathbf{y}  ^{(2)} _i \in \mathbb{R} ^q $, of sizes $N _1 $ and $N _2 $, respectively. The first pair  will be used for training and the second one for testing. Suppose that all the elements in these random samples satisfy hypothesis  {\bf (A1)} and that the ridge residuals $ \left\{\boldsymbol{\varepsilon}  ^{(1)} _{ \lambda , i} \right\} _{i \in \left\{ 1, \dots, N _1 \right\} }$ associated to the training pair are   conditionally normal and homoscedastic in the sense of Lemma \ref{Lemma 2}, that is, their corresponding conditional covariance matrices are constant and equal to some matrix $\Sigma _{  \boldsymbol{\varepsilon}   | {\mathbf{x}  } } ^ {\lambda, (1)}\in \Bbb S _q $. 
Consider now the random matrices $X _1 \in \mathbb{M} _{p, N _1 }$ and $Y _1  \in \mathbb{M} _{q, N _1 }$ obtained by horizontal concatenation of the elements in the training samples. Under these hypotheses, the conditional total testing error introduced in~\eqref{MSE testing} can be written as:
\begin{align}
\label{MSE test}
{\rm MSE} ^\lambda _{{\rm testing}|X_1 } &= {\rm trace} \left( \Sigma  ^{(2)} _{\mathbf{y} }  + \boldsymbol{\mu} ^{(2)} _{\mathbf{y} } \boldsymbol{\mu} _{\mathbf{y}  }^ {(2)\top}   \right) - 2\, {\rm trace}\left( \left(\Sigma  ^{(2)}_{\mathbf{x}   \mathbf{y}  }  + \boldsymbol{\mu} _{ \mathbf{x}   }^{(2)}\boldsymbol{\mu} _{\mathbf{y} } ^{(2)\top} \right) M_{ \widehat{B} _\lambda}^\top \right) \nonumber\\
&+{\rm trace} \left[ \left( \Sigma^{(2)}  _{\mathbf{x}  }  + \boldsymbol{\mu}^{(2)} _{\mathbf{x}  } \boldsymbol{\mu} _{\mathbf{x}  }^ {(2)\top} \right) \left({\rm trace} \left(\Sigma ^{\lambda, (1)}_{ \boldsymbol{\varepsilon} | \mathbf{x} }\right)  Z _\lambda   R _\lambda   + M_{ \widehat{B} _\lambda}M ^{ \top}_{ \widehat{B} _\lambda} \right)\right],
\end{align}
where 
\begin{align}
Z _\lambda  &:=R _\lambda X _1 A_{N_1} X _1 ^\top ,\label{Z test}\\
R _\lambda  &:=\left(X _1 A_{N_1} X _1 ^\top + \lambda N _1 \mathbb{I}_p \right) ^{-1},\label{R test} \\
M _{ \widehat{B} _\lambda}&:= B _\lambda - \lambda N _1 R _\lambda B _\lambda + R _\lambda X _1 A _{N_1} M _{ E_\lambda | X} ^{(1)\top}, \label{M test}\\
\Sigma  ^{(2)} _{\mathbf{y} }   &= {\rm Cov} ( \mathbf{y} ^{(2)} ,\mathbf{y} ^{(2)} ),\label{Sigma y test}\\
\Sigma  ^{(2)} _{\mathbf{x} }   &= {\rm Cov} (\mathbf{x} ^{(2)}  ,\mathbf{x} ^{(2)}  ),\label{Sigma x test}\\
\Sigma  ^{(2)} _{\mathbf{x} \mathbf{y} }  &= {\rm Cov} ( \mathbf{x} ^{(2)}  ,\mathbf{y} ^{(2)} ).\label{Sigma xy test}
\end{align}
with 
\begin{align}
\label{}
A _{N_1} &:= \mathbb{I} _{N_1} - \dfrac{1}{N _1 } \mathbf{i} _{N _1 }\mathbf{i} _{N _1 } ^\top.
\end{align}
\end{theorem}

\noindent$\blacktriangleright$ {\bf Example~A (continued).} As we already pointed out when working with the training error, the homoscedasticity hypothesis is satisfied in the setup  of Example~A with $\Sigma ^{\lambda, (1)}_{ \boldsymbol{\varepsilon} | \mathbf{x} } = \Sigma ^{(1)}_{ \boldsymbol{\varepsilon} }$ and  the properties of the estimator $\widehat{B} _\lambda $ can be formulated in this case in terms of the matrix $B$. These observations lead us to the following corollary of Theorem \ref{Theorem testing error} that provides an explicit formula for the generalization error in this setup. 

\begin{corollary} 
\label{Corollary testing error Example A} 
In the conditions of Example~A  the conditional total testing error is determined by the following expression:
\begin{align}
\label{MSE test example A}
{\rm MSE} ^\lambda _{{\rm testing}|X_1 } &= {\rm trace} \left( \Sigma  ^{(2)} _{\mathbf{y} }  + \boldsymbol{\mu} ^{(2)} _{\mathbf{y} } \boldsymbol{\mu} _{\mathbf{y}  }^ {(2)\top}   \right) - 2\, {\rm trace}\left( \left(\Sigma  ^{(2)}_{\mathbf{x}   \mathbf{y}  }  + \boldsymbol{\mu} _{ \mathbf{x}   }^{(2)}\boldsymbol{\mu} _{\mathbf{y} } ^{(2)\top} \right) B^\top Z _\lambda ^\top \right) \nonumber\\
&+{\rm trace} \left[ \left( \Sigma^{(2)}  _{\mathbf{x}  }  + \boldsymbol{\mu}^{(2)} _{\mathbf{x}  } \boldsymbol{\mu} _{\mathbf{x}  }^ {(2)\top} \right) \left({\rm trace} (\Sigma ^{ (1)}_{ \boldsymbol{\varepsilon}  })  Z _\lambda   R _\lambda   + Z _\lambda B B ^\top Z_\lambda ^\top \right)\right],
\end{align}
with  $R _\lambda$, $\Sigma  ^{(2)} _{\mathbf{y} } $, $\Sigma  ^{(2)} _{\mathbf{x} }$, and $\Sigma  ^{(2)} _{\mathbf{x} \mathbf{y} }$ as in \eqref{R test}, \eqref{Sigma y test}, \eqref{Sigma x test}, and \eqref{Sigma xy test}, respectively. $\blacktriangleleft$,
\end{corollary}

\subsection{Numerical illustration}
\label{Numerical illustration}
The goal of this section is showing how to compute in an explicit example the total training and testing error formulas that we provided in Theorems~\ref{Theorem 2} and~\ref{Theorem testing error} and to show how they can be used at the time of deciding on a modeling architecture that minimizes overfitting and generalization errors. On other words, we will see how the theoretical results that we just provided can help in finding the optimal trade-off between modeling complexity and out-of-sample performance. 

\medskip

\noindent {\bf The underlying model.} Consider a particular instance of Example B where the dependent variable ${y} $ is generated by the additive stochastic perturbation of a deterministic model of the form
\begin{equation}
\label{exampleB model2}
{y} = f(\xi ) + {\varepsilon}, 
\end{equation}
with
\begin{equation}
\label{f xi}
f(\xi ) = e^ \xi  - k, \enspace {\rm with} \enspace 
k = \dfrac{1}{2}( e - e^{-1}).
\end{equation}
The term ${\varepsilon} \sim {\rm N} ( 0 , \sigma_{ \varepsilon }^2 )$ in the nonlinear model \eqref{exampleB model2}  represents the disturbance term. Additionally, we assume that  $\xi $ is uniformly distributed in the interval $[-1,1] $, that is,  ${ \xi }\sim \mathcal{U} ( -1 , 1 )$ and hence has  probability density function $g_\xi(z) = \frac{1}{2} I_{[-1,1]}(z) $. As it was assumed in the general conditions of Example B, we suppose that $\varepsilon $ and $\xi $ are independent random variables and we require that ${\rm E} \left[ f( \xi )\right] = 0$,  which is automatically guaranteed by the choice of $k$ in \eqref{f xi}.  

\medskip

\noindent  {\bf A polynomial approximation and the corresponding linear regression model.}
We  now construct a polynomial approximation of  \eqref{exampleB model2} and, as in Example B, we formulate this procedure as a linear univariate regularized regression model of the form
\begin{equation}
\label{approximated model example B1}
y = {{\bf B}} _\lambda ^\top \mathbf{x}  + \varepsilon _\lambda, 
\end{equation}
where the covariates $\mathbf{x} $ are adequately standardized powers of increasing order of the random variable $\xi$. More specifically, we set:  
\begin{equation}
\label{x example B2}
\mathbf{x} := \left( \widetilde{\xi} , \widetilde{\xi^2}, \dots, \widetilde{ \xi^p } \right) ^\top, \enspace {\rm with }\enspace x _i = \widetilde{ \xi ^i} := \dfrac{ \xi ^i - {\rm E} \left[ \xi ^i\right] }{ \sigma (\xi ^i)}, \enspace i \in \left\{ 1, \dots,  p\right\},
\end{equation}
Notice that the polynomials that we are using in the construction of the explanatory variables $\mathbf{x}  $ are not orthogonal and, moreover, produce covariance matrices $\Sigma _{\bf x} $ whose condition numbers increase rapidly with the degree $p$. This feature makes pertinent in this context the use of the ridge regularization.

Given that  ${ \xi }\sim \mathcal{U} ( -1 , 1 )$, we have: 
\begin{equation}
\label{exp and std}
 {\rm E} \left[ \xi ^i\right] = \dfrac{(-1)^i + 1}{2(i+1)}, \enspace  {\rm and} \enspace \sigma \left( \xi ^i\right) = \left(\dfrac{1}{2i+1} -  \dfrac{(-1)^i + 1}{2(i+1)^2}\right)^{1/2}. 
\end{equation}
Once a regularization strength $\lambda \in \mathbb{R} ^+$ has been fixed, the corresponding regularized regression vector  ${\bf B} _\lambda $  is given by \eqref{B lambda}, that is, ${\bf B} _\lambda := \left(\Sigma _{\bf x}  + \lambda \mathbb{I}_p \right) ^{-1}  {\mathbf{\Sigma}} _{{\bf x}y} $. In this relation $\Sigma _{\bf x} \in \Bbb S _p $ and $ {\mathbf{\Sigma}} _{{\bf x}y} \in \mathbb{R}^p $ can be explicitly computed using the fact that ${ \xi }\sim \mathcal{U} ( -1 , 1 )$. Indeed, for $i, j\in \left\{ 1, \dots, p \right\}$  we have: 
\begin{align}
\label{sigma x example B2}
(\Sigma _{\bf x})_{i,j}=&\dfrac{1}{\sigma (\xi ^i)\sigma (\xi ^j)}\left(  {\rm E} \left[ \xi ^{i+j}\right] - {\rm E} \left[ \xi ^i\right]{\rm E} \left[ \xi ^j\right]\right) ,\\
(\mathbf{\Sigma} _{{\bf x}y})_{i}=& \dfrac{1}{2\sigma (\xi ^i)} \Big( (-1)^i \sum^{i}_{s = 0} \dfrac{i!}{s!} ((-1)^s e - e ^{-1}) - 2 k {\rm E} \left[ \xi ^i\right]\Big), \label{sigma xy example B2} 
\end{align}
with  ${\rm E} \left[ \xi ^i\right] $ and $\sigma ( \xi^i)$ as in \eqref{exp and std}. In order to complete the model specification in \eqref{approximated model example B1} we determine the regression residuals $ \varepsilon _\lambda $ that are given by
\begin{equation}
\label{residuals lambda example B2}
\varepsilon _\lambda  = e^ \xi -k - {\bf B}_\lambda ^\top \mathbf{x} + \varepsilon.
\end{equation}
The general results in Example B provide the following expressions for the first and the second unconditional moments of $\varepsilon _\lambda $
\begin{align}
\label{mu eps lambda example B2}
\mu _{\varepsilon _\lambda } &= 0,\\
\sigma ^2 _{\varepsilon_\lambda } &= \sigma ^2 _{y} - {\bf B}_\lambda ^\top \mathbf{ \Sigma }_{\mathbf{x} y} - \mathbf{ \Sigma }_{\mathbf{x} y} ^\top {\bf B}_\lambda + {\bf B}_\lambda ^\top \Sigma _{ \mathbf{x} } {\bf B}_\lambda, \label{sigma eps lambda example B2} 
\end{align}
where
\begin{equation}
\label{sigma y ExampleB2}
\sigma ^2 _{y} = \dfrac{1}{2}(1-e^{-2}) + \sigma _\varepsilon ^2.
\end{equation}

\noindent  {\bf The conditional training and testing errors.} We now use the formulas in Theorems~\ref{Theorem 2} and~\ref{Theorem testing error} in order to asses the performance of the 
approximating linear regression model \eqref{approximated model example B1} as a function of its complexity or, more specifically, its ability to reproduce the underlying nonlinear relation between $\xi $ and $y$. We will carry this out in two different situations that put the accent in the overfitting and in the lack of generalization power that the model may incur when the selected number of covariates is too high. These features will be detected by studying the evolution of the relative values of the conditional training and testing errors as a function of the model parsimony.

Regarding the conditional total training error, as we  explained earlier, the training sample is generated using the underlying model in \eqref{exampleB model2}-\eqref{f xi} with $\xi \sim \mathcal{U}(-1,1)$. The sample length is denoted by $N _1 $. In order to evaluate the conditional total training error in Theorem~\ref{Theorem 2}, the following expressions for  
the conditional first and the second moments of the ridge residuals are required
\begin{align}
\label{mu eps lambda cond x example B1}
\mu _{\varepsilon _\lambda | \mathbf{x} } &= e^ \xi -k - {\bf B} _\lambda ^\top\mathbf{x} ,\\
\sigma ^2 _{\varepsilon_\lambda  | \mathbf{x} } &= \sigma _\varepsilon ^2. \label{sigma eps lambda cond x example B1} 
\end{align}

As to the conditional total testing error,  we consider a general case in which the random testing sample $({y^{(2)}}, {\bf x}^{(2)})$ of length $N _2 $ is generated using the underlying model \eqref{exampleB model2}-\eqref{f xi} but, this time around, we allow the support of the random variable used to generate the dependent variable $y^{(2)} $ to be an arbitrary closed interval, that is,  
\begin{equation}
\label{true model testing}
y^{(2)}=e^{\zeta } - k  + \varepsilon,
\end{equation}
where $k$ is given in \eqref{f xi}, $\varepsilon \sim {\rm N} (0, \sigma _\varepsilon ^2 )$, and $\zeta   \sim \mathcal{U}(a, b)$ with  $a,b \in \mathbb{R} $ such that $a<b $. At the same time, the covariates sample ${\bf x}^{(2)}$ is constructed as in \eqref{x example B2}, that is,
 \begin{equation}
\label{x example B2 test}
\mathbf{x}^{(2)} := \left( \widetilde{\zeta } , \widetilde{\zeta^2}, \dots, \widetilde{ \zeta^p } \right) ^\top, \enspace {\rm with }\enspace {(\mathbf{x}^{(2)})} _i = \widetilde{ \zeta ^i} := \dfrac{ \zeta ^i - {\rm E} \left[ \xi ^i\right] }{ \sigma (\xi ^i)}, \enspace i \in \left\{ 1, \dots,  p\right\},
\end{equation}
where $ {\rm E} \left[ \xi ^i\right]$ and $\sigma \left( \xi ^i\right)$ for ${ \xi }\sim \mathcal{U} ( -1 , 1 )$ are defined in \eqref{exp and std}. 
The sample matrix $X^{(2)} \in \mathbb{M} _{p, N_2}$ and vector  ${\bf y}^{(2)\top} \in \mathbb{R} ^{N_2}$ are obtained by horizontal concatenation of the $N_2$ realizations of $({y^{(2)}}, {\bf x}^{(2)})$.
In order to evaluate the conditional total testing error in \eqref{MSE test}, we need  the first and second order moments of the random testing sample. First, we notice that  since $\xi^{(2)}  \sim \mathcal{U}(a, b)$, we can conclude that:
\begin{align}
\label{}
(\boldsymbol{\mu} _{\mathbf{x} }^{(2)})_i = \dfrac{1}{\sigma \left( \xi^i\right)} \left( {\rm E} \left[ \zeta ^i\right] - {\rm E} \left[ \xi  ^i\right]\right) , \enspace i \in \left\{ 1, \dots, p \right\},
\end{align}
where 
\begin{equation}
\label{exp zeta}
{\rm E} \left[ \zeta ^i\right] = \dfrac{b^{i+1}-a^{i+1}}{(i+1)(b-a)}
\end{equation}
and where for each $i \in \left\{ 1, \dots, p \right\}$,  $ {\rm E} \left[ \xi ^i\right]$ and $\sigma \left( \xi ^i\right)$ are given by \eqref{exp and std}.
Additionally, the first order moment of  $y^{(2)}$ is given by
\begin{equation}
\label{mu y 2}
\mu _{y}^{(2)} = \dfrac{1}{b-a} (e^b-e^a)-k.
\end{equation}
The  second order moments of the testing sample are:
 \begin{align}
\label{sigma x example B test}
(\Sigma _{\bf x}^{(2)})_{i,j}=& \dfrac{1}{(b-a)\sigma (\xi ^i)\sigma (\xi ^j)}\int_{a}^{b} \left(  z^i - {\rm E} \left[ \xi ^i\right]\right) \left( z^j - {\rm E} \left[ \xi ^j\right]\right)  dz - (\boldsymbol{\mu} _{\mathbf{x} }^{(2)})_i(\boldsymbol{\mu} _{\mathbf{x} }^{(2)})_j, \enspace i, j\in \left\{ 1, \dots, p \right\}, \\
(\mathbf{\Sigma} _{{\bf x}y}^{(2)})_{i}=& \dfrac{1}{(b-a)\sigma (\xi ^i)}\int_{a}^{b} \left(  z^i - {\rm E} \left[ \xi ^i\right]\right)(e^z -k) dz - (\boldsymbol{\mu} _{\mathbf{x} }^{(2)})_i \mu _y^{(2)}, \enspace i\in \left\{ 1, \dots, p \right\}, \label{sigma xy example B test} \\
(\sigma ^2_y)^{(2)} =& \dfrac{1}{2 (b-a)} (e^{2b} - e^{2a}) - \dfrac{1}{(b-a)^2} (e^b - e^a)^2 + \sigma_{ \varepsilon }^2.
\end{align}
These relations allow for the explicit evaluation of the conditional total testing error that we now numerically study for two particular choices for the values of $a$ and $b$ that determine the support of the distribution of $ \zeta $.

\medskip

\begin{centering}
 \begin{figure}[!h]
 \includegraphics[width=1.1\textwidth]{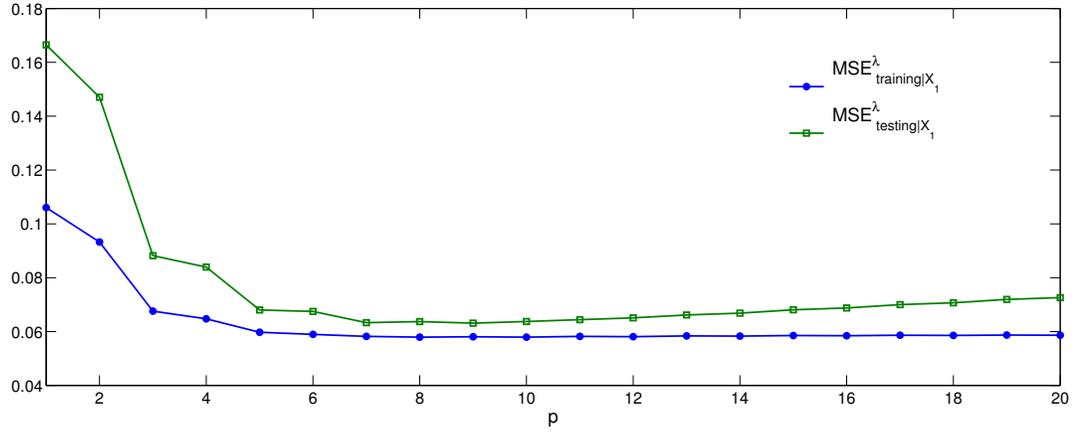}
 \caption{Conditional total training and  testing errors  committed by the regularized ($ \lambda =0.9$) polynomial approximation  of the  nonlinear underlying model \eqref{exampleB model2}-\eqref{f xi}  as a function of the number of covariates $p$ or, equivalently, the degree of the polynomial approximation. The lengths of the training and testing samples are $N _1 =20$ and $N _2 = 20$, respectively, and $\zeta, \xi  \sim \mathcal{U}(-1,1)$. }
  \label{Figure 1} 
 \end{figure}
 \end{centering}

\noindent {\bf Numerical illustration. Case I. Overfitting.} Consider first the case in which $[a,b]=[-1,1]$ and hence $ \zeta = \xi$. This means that the testing sample is generated in the same way as the training one. In this setup, an increase in the number of covariates $p$ amounts to an increase in the degree of the polynomial that we are using to fit the sample. When this degree is too high, the fit may lead to a match with the noise in the sample rather than with the underlying deterministic signal that we are trying to model. Figure~\ref{Figure 1} shows the evolution of the conditional training and total errors for this case (for a given training covariates sample $X _1$ and a fixed regularization strength reported in the caption). As we anticipated, the conditional total training error is a decreasing function with $p$ but, in turn, the generalization error decreases only up to a point and when overfitting to the training sample starts to occur, we observe a monotonous increase in the testing error. In this case, this turning point takes place for $p=9$, a value that seems to offer the optimal tradeoff between the quality of training and out-of-sample performance. Another way to state this observation is that, with this noise level, the best  polynomial approximation (in the mean square sense) of the the exponential function in the interval $[-1,1]$ is obtained when using polynomials of degree  $9$. 	
\medskip

\begin{figure}[!h]
 \includegraphics[width=1\textwidth]{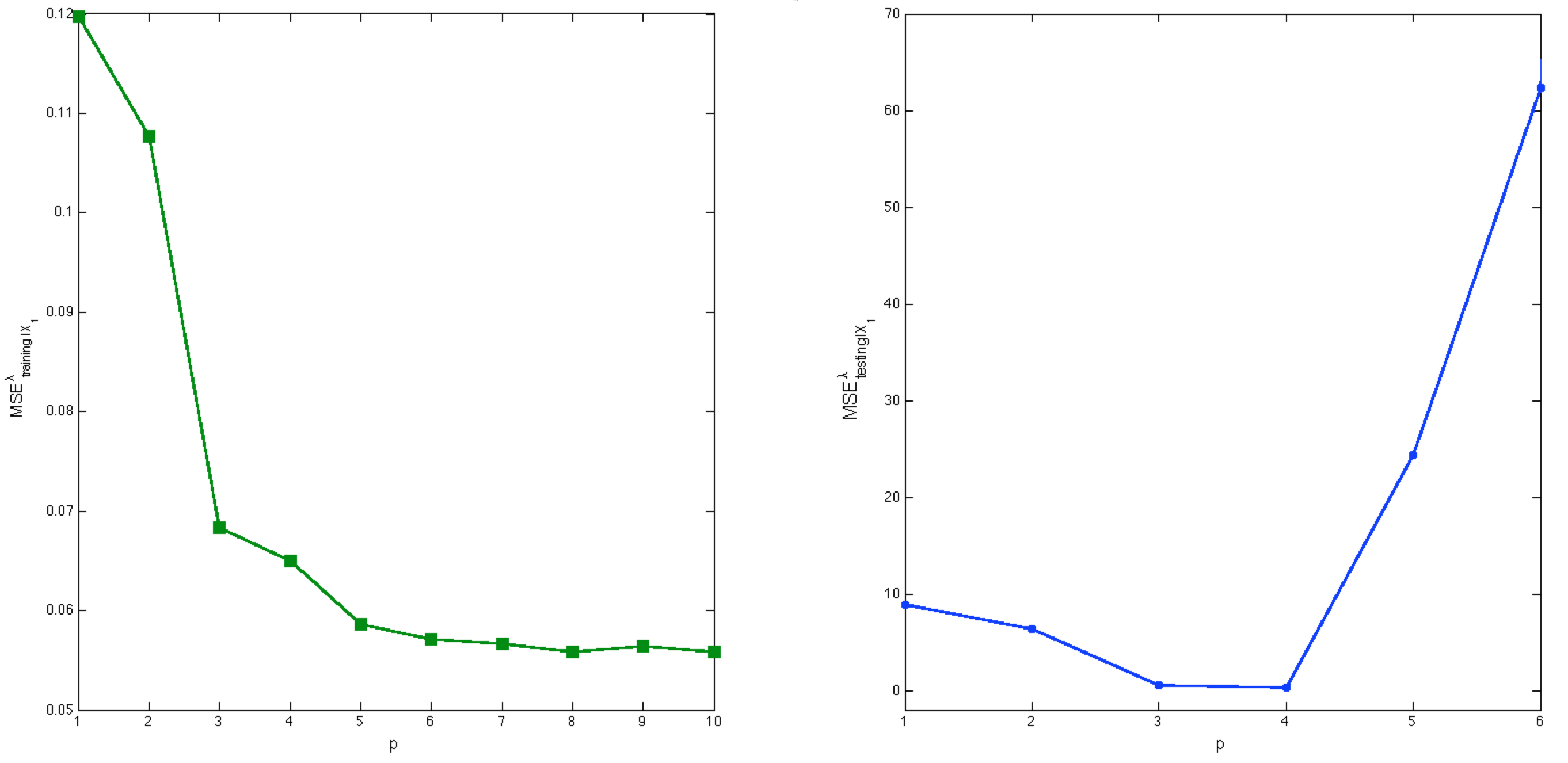}\\
 \caption{Conditional total training and  testing errors  committed by the regularized ($ \lambda =1$) polynomial approximation  of the  nonlinear underlying model \eqref{exampleB model2}-\eqref{f xi}  as a function of the number of covariates $p$ or, equivalently, the degree of the polynomial approximation.
The lengths of the training and testing samples are $N _1 =30$ and $N _2 = 40$, respectively. The random variables $\xi  \sim \mathcal{U}(-1,1)$ and  $\zeta \sim \mathcal{U}(1,2)$ have been used in the dependent variables generating model.}
 \label{Figure 2} 
 \end{figure}

\noindent {\bf Numerical illustration. Case II. Generalization performance.}  In this case we allow for the support of distribution for $\zeta $ used to generate the testing sample to be different from the one that defines the construction of the training sample.  Indeed, we use $\xi \sim \mathcal{U}(-1,1)$ and $\zeta  \sim \mathcal{U}(1,2)$ and hence the training and testing are carried out in the different intervals where $\xi $ and $\zeta $ are defined. Equivalently, in this situation the testing will be measuring the ability of the approximate model to reckon the values of the underlying model in an interval different from the one in which the training has taken place; that is why we explicitly talk in this case of generalization performance. Figure~\ref{Figure 2} shows the evolution of the conditional training and total errors for this case (for a given training covariates sample $X _1$ and a fixed regularization strength reported in the caption). Again, the conditional total training error is a decreasing function of $p$. As to the testing error in this case, its absolute value is much higher than in Case I and its minimum is already attained for $p=4$, that is, in this case a fourth order polynomial provides the best tradeoff in the modeling of the exponential function. 
\section{Appendices}
\label{Appendices}

\subsection{Proof of Theorem \ref{Theorem 1}}
\label{Proof of Theorem 1} 
We first notice that by  definition of $R _\lambda $ in \eqref{R lambda repeat}, namely $R _\lambda =(XA _N X ^\top +\lambda N \mathbb{I}_p )^{-1}$, the relation \eqref{B lambda random sample est} can be rewritten as $\widehat{B }_\lambda= R _\lambda X A_N Y ^\top$. In this expression we use that $Y=B _\lambda ^\top X + E_\lambda $ and that, by the same definition of $R _\lambda $ in \eqref{R lambda repeat},  $XAX ^\top = R _\lambda ^{-1} -\lambda N \mathbb{I}_p $. Taking  both observations into account, we obtain that $\widehat{B }_\lambda= R _\lambda X A_N Y ^\top = R _\lambda X A_N  ( X ^\top B_\lambda + E _\lambda ^\top) = R _\lambda X A_N  X ^\top B_\lambda + R _\lambda X A_N E _\lambda ^\top = B _\lambda - \lambda N R _\lambda B _\lambda + R _\lambda X A _N  E _\lambda ^\top$ or, equivalently,  
\begin{equation}
\label{}
\widehat{B} _\lambda - B _\lambda + \lambda N R _\lambda B_\lambda = R _\lambda X A _N  E_\lambda ^\top.
\end{equation}
We now point out that the conditional homoscedasticity hypothesis on the residuals implies by Corollary~\ref{Corollary Lemma 2} that  $E _{\lambda} |X \sim {\rm MN} ( M_{ E _\lambda | X}, \Sigma _{  \boldsymbol{\varepsilon}  | \mathbf{x} } ^ \lambda, \mathbb{I}_N )$ and hence by Theorem 2.3.10 in \cite{bookMatrixDistributions2000} we have that 
\begin{equation}
\label{}
(\widehat{B} _\lambda - B _\lambda + \lambda N R _\lambda B_\lambda)|X \sim {\rm MN} ( R _\lambda X A _N M_{ E _\lambda | X} ^\top, R_\lambda X A _N A _N ^\top X ^\top R_\lambda , \Sigma _{  \boldsymbol{\varepsilon}  | \mathbf{x} } ^ \lambda).
\end{equation}
Since $A _N A _N ^\top = A_N $, the statement in \eqref{B lambda hat minus B lambda} follows. $\blacksquare$

\subsection{Proof of Lemma \ref{Lemma 1}}
\label{Proof of Lemma 1} 
\noindent  {\bf Part (i)}. By \eqref{B lambda random sample est} we have that $\widehat{B} _\lambda :=(XA_NX^\top + \lambda N \mathbb{I}_p ) ^{-1}  XA_NY^\top$. At the same time the relation \eqref{B lambda0 random sample est} yields  $XA_NY^\top = (XA_N X ^\top ) \widehat{B}$, which in \eqref{B lambda random sample est} gives $\widehat{B} _\lambda :=(XA_NX^\top + \lambda N \mathbb{I}_p ) ^{-1}  (XA_N X ^\top ) \widehat{B}$. The last expression is equivalent to ~\eqref{B lambda ZB}, as required. 

\noindent{\bf Part (ii)}. First, in order to establish \eqref{Z1}, we notice that from the definition of $R _\lambda $ in  \eqref{R lambda}, it follows that $XA_N X ^\top = R _{\lambda} ^{-1} - \lambda N \mathbb{I} _p$ and substituting this relation into \eqref{Z lambda} in   {\bf (i)}   yields 
\begin{align*}
Z _\lambda = R _{\lambda} XA_N X ^\top = R _{\lambda} \left( R _{\lambda} ^{-1} - \lambda N \mathbb{I} _p \right) = \mathbb{I}_p - \lambda N R _\lambda,
\end{align*}
as required. Second, we show that \eqref{Z2} holds. Rewriting the relation \eqref{Z lambda} using the  definition of $R _\lambda $ in  \eqref{R lambda}, we obtain that 
\begin{align*}
Z _\lambda =\left((XA_N X ^\top) ^{-1} XA_N X ^\top + \lambda N (XA_N X ^\top) ^{-1} \right) ^{-1} = \left( \mathbb{I} _p + \lambda N (XA_N X ^\top) ^{-1}  \right) ^{-1},
\end{align*}
we establish that it is equivalent to  \eqref{Z2}, as required. $\blacksquare$

\subsection{Proof of Corollary \ref{Corollary Theorem 1}}
\label{Proof of Corollary of Theorem 1} 
As $\widehat{B} _\lambda - B = (\widehat{B}_\lambda - B_\lambda ) +  ({B}_\lambda - B )$, then from \eqref{B lambda hat minus B lambda} in Theorem \ref{Theorem 1}  follows that 
\begin{align}
\label{B lambda minus B proof} 
(\widehat{B} _\lambda - B )|X \sim {\rm MN} (B _\lambda - B-\lambda N R _\lambda B_\lambda + R_\lambda X A _N   M_{ E _\lambda | X} ^\top, R_\lambda X A _N  X ^\top R_\lambda ,\Sigma _{  \boldsymbol{\varepsilon}  | \mathbf{x} } ^ \lambda).
\end{align}
 First, notice that by \eqref{mu eps lambda cond x example} and by \eqref{mu E lambda cond X} we can conclude that  $M_{ E _\lambda | X} ^\top = X ^\top (B - B _\lambda )$. Consequently, 
 \begin{align}
 \label{mean proof B lambda hat minus B} 
B _\lambda - B -\lambda N R _\lambda B_\lambda + R _\lambda X A_N X ^\top \left(B - B _\lambda \right)&= B _\lambda - B  - \lambda N R _\lambda B _\lambda + R _\lambda (R _\lambda ^{-1} -\lambda N \mathbb{I}_p ) \left( B - B _\lambda \right) \nonumber \\
 &= B _\lambda - B - \lambda N R _\lambda B_\lambda + B - B _\lambda -\lambda N R _\lambda B +\lambda N R _\lambda B _\lambda \nonumber \\
 &= - \lambda N R_\lambda B,
\end{align}
 where we used the definition of $R_\lambda $ in \eqref{R lambda repeat}. Second, we rewrite the first scale matrix in \eqref{B lambda minus B proof} as follows
 \begin{align}
\label{RXAR proof}
R _\lambda XA_N X ^\top R _\lambda  &= R_\lambda (R _\lambda^{-1} - \lambda N \mathbb{I}_p ) R _\lambda  = \left( \mathbb{I} _p - \lambda N R _\lambda \right) R_\lambda  = Z _\lambda R_\lambda.
\end{align}
At the same time the second scale matrix in \eqref{B lambda minus B proof} in the conditions of the statement is equal  to $\Sigma _{\boldsymbol{\varepsilon} }$ and hence together with  \eqref{mean proof B lambda hat minus B} and \eqref{RXAR proof} this yields \eqref{B lambda minus B our notation}.

Now, in order to show \eqref{B lambda minus B}, we first use the definition of $Z _\lambda $ in \eqref{Z1} and rewrite \eqref{RXAR proof} as
\begin{align}
 R _\lambda XA_N X ^\top R _\lambda&= Z _\lambda R_\lambda = \dfrac{1}{ \lambda N} Z_\lambda \left( \mathbb{I} _p - Z _\lambda \right) = \dfrac{1}{ \lambda N} \left( Z _\lambda - Z _\lambda ^2 \right)=\dfrac{1}{\lambda N } Z_\lambda\left( Z _\lambda ^{-1} - \mathbb{I} _p\right) Z _\lambda.
\end{align}
As by hypothesis $X A _N X ^\top$ is invertible, then by \eqref{Z2} it holds that $(X A _N X ^\top )^{-1} = \dfrac{1}{ \lambda N}\left( Z _\lambda ^{-1} - \mathbb{I}_ p \right) $ and the expression \eqref{RXAR proof} becomes
 \begin{align}
\label{RXAR proof2}
R _\lambda XA_N X ^\top R _\lambda = Z _\lambda(X A _N X ^\top )^{-1} Z _\lambda.
\end{align}
Consequently, the relations \eqref{B lambda minus B proof} and \eqref{mean proof B lambda hat minus B} together with \eqref{RXAR proof2} guarantee \eqref{B lambda minus B}, as required. $\blacksquare$

\subsection{Proof of Theorem \ref{Theorem 2}}
\label{Proof of Theorem 2}  We start by using the definition of the conditional total training error provided in \eqref{MSE training}. First, we define $D_\lambda := \widehat{B} _\lambda  - B _\lambda $ and subtract and add    $ B _\lambda $ to $\widehat{B} _\lambda $ in \eqref{MSE training} which  results in 
\begin{align}
\label{proof mse training}
{\rm MSE}_{{\rm training}|X}^\lambda &= \dfrac{1}{N} {\rm trace}\left( {\rm E} _X\left[ (Y - (\widehat{B} _\lambda - B _\lambda + B _\lambda ) ^\top X) ^\top (Y - (\widehat{B} _\lambda - B _\lambda + B _\lambda )X)\right] \right) \nonumber \\
&= \dfrac{1}{N} {\rm trace}\left( {\rm E} _X\left[ (E _\lambda  - (\widehat{B} _\lambda - B _\lambda  ) ^\top X) ^\top (E _\lambda  - (\widehat{B} _\lambda - B _\lambda  ) ^\top X)\right] \right)\nonumber \\
&= \dfrac{1}{N} {\rm trace}\left( {\rm E} _X\left[ (E _\lambda  - D _\lambda  ^\top X) ^\top (E _\lambda  - D _\lambda  ^\top X)\right] \right)= \dfrac{1}{N} {\rm trace} ({\rm E} _X\left[ E _\lambda^\top  E_\lambda \right] )\nonumber \\
& + \dfrac{1}{N} {\rm trace} (X ^\top {\rm E}_X \left[ D_\lambda D _\lambda ^\top\right] X ) - \dfrac{2}{N} {\rm trace}\left( X^\top {\rm E} _X\left[  D _\lambda E _\lambda \right] \right).
\end{align}
 We now study separately the three summands in \eqref{proof mse training}. The first one can be computed in a straightforward way. Indeed, by {\bf (i)} in Lemma \ref{Lemma 2} we can immediately write
 \begin{equation}
\label{proof trace ee x}
 \dfrac{1}{N} {\rm trace} ({\rm E} _X\left[ E _\lambda ^\top E _\lambda \right] ) = {\rm trace} (\Sigma _{  \boldsymbol{\varepsilon}  | \mathbf{x} } ^ \lambda ) + \dfrac{1}{N} {\rm trace} (M _{E_\lambda |X} ^\top M _{E_\lambda |X}).
\end{equation}
As for the second summand, we notice first that, in the hypotheses of the theorem,  the properties of the estimator $\widehat{B} _\lambda$ are as provided in Theorem~\ref{Theorem 1}, that is, 
\begin{equation}
\label{B lambda hat minus B lambda repeat}
D_\lambda|X \sim {\rm MN} (M _{D _\lambda }, R_\lambda X A _N  X ^\top R_\lambda ,\Sigma _{  \boldsymbol{\varepsilon}  | \mathbf{x} } ^ \lambda), 
\end{equation}
with 
\begin{equation}
\label{mu M proof}
M _{D _\lambda }:=-\lambda N R _\lambda B_\lambda + R_\lambda X A _N   M_{ E _\lambda | X} ^\top
\end{equation}
and
\begin{equation}
\label{R lambda repeat proof}
R _\lambda :=(XA _N X ^\top +\lambda N \mathbb{I}_p )^{-1}.
\end{equation}
At the same time, by Theorem 2.3.10 in \cite{bookMatrixDistributions2000} (see also Lemma 6.3 in \cite{RC3}) we have that $D _\lambda$ satisfies the following relation
\begin{align}
\label{covRowproperty} 
{\rm E}_X \left[ D _\lambda D_\lambda ^\top\right] & =  {\rm trace}(\Sigma _{  \boldsymbol{\varepsilon}  | \mathbf{x} } ^ \lambda)R_\lambda X A _N  X ^\top R_\lambda + M _{D _\lambda }M _{D _\lambda }^\top.
\end{align}
Using this expression, we see that the second summand in \eqref{proof mse training} can be written as
\begin{align}
\label{proof trace mlml x}
\dfrac{1}{N} {\rm trace}\left( X ^\top {\rm E}_X \left[ D _\lambda D_\lambda ^\top\right] X  \right) &=  \dfrac{1}{N} {\rm trace}\left( {\rm E}_X \left[ D _\lambda D_\lambda ^\top\right] XX ^\top   \right) \nonumber\\
&= \dfrac{1}{N} {\rm trace}\Bigg[ \left( {\rm trace}( \Sigma _{  \boldsymbol{\varepsilon}  | \mathbf{x} } ^ \lambda) R _\lambda X A _N X ^\top R_\lambda + M _{D _\lambda }M _{D _\lambda }^\top \right) X X ^\top\Bigg]\nonumber\\
&=\dfrac{1}{N} {\rm trace}( \Sigma _{  \boldsymbol{\varepsilon}  | \mathbf{x} } ^ \lambda)Z _\lambda R _\lambda XX^\top +\dfrac{1}{N}{\rm trace}(M _{D _\lambda }M _{D _\lambda }^\top  X X ^\top),
\end{align}
where \begin{equation}
\label{Z lambda repeat}
Z _\lambda := R _\lambda X A _N X^\top.
\end{equation}
Finally,  we can rewrite the third summand in  \eqref{proof mse training} by using the definition of the ridge regression matrix estimator $\widehat{B} _\lambda $ in \eqref{B lambda random sample est} and using  $R_\lambda $ in \eqref{R lambda repeat proof}: 
\begin{align*}
X ^\top {\rm E}_X \left[  D _\lambda E _\lambda \right] = & X ^\top {\rm E}_X \left[  \left(\widehat{B} _\lambda -B _\lambda \right) E_\lambda  \right] = X ^\top {\rm E} _X\left[ \widehat {B} _\lambda  E_\lambda  \right] - X ^\top B_\lambda {\rm E} _X\left[  E _\lambda  \right]\\
=&X^\top R _\lambda X A _N  {\rm E}_X \left[ Y^\top E _\lambda  \right]  - X^\top B _\lambda {\rm E} _X\left[ E_\lambda  \right] =X  ^\top  R _\lambda X_N  A _N {\rm E} _X\left[ X ^\top B _\lambda E _\lambda + E_\lambda ^\top E _\lambda \right] \\ 
&- X^\top B _\lambda {\rm E} _X\left[ E_\lambda  \right]  =X ^\top R _\lambda X A_N (X ^\top B _\lambda {\rm E} _X\left[ E_\lambda  \right] + {\rm E} _X\left[ E_\lambda^\top E_\lambda  \right])- X^\top B _\lambda {\rm E} _X\left[ E_\lambda  \right].
\end{align*}
In this expression we now use the part {\bf (i)} in Lemma \ref{Lemma 2} and hence conclude that
\begin{align}
\label{proof E xmle x}
-\dfrac{2}{N}{\rm trace} \left({\rm E}_X \left[ X ^\top D _\lambda E _\lambda  \right] \right)=&-\dfrac{2}{N}{\rm trace}\Big( X ^\top R _\lambda X A_N  (X ^\top B _\lambda M _{E_\lambda |X}  + {\rm trace}(  \Sigma _{  \boldsymbol{\varepsilon}  | \mathbf{x} } ^ \lambda ) \mathbb{I} _N + M _{E_\lambda |X}  ^\top M _{E_\lambda |X}  )  \nonumber\\
&- X^\top B _\lambda M _{E_\lambda |X} \Big) = -\dfrac{2}{N}\Big({\rm trace}( \Sigma _{  \boldsymbol{\varepsilon}  | \mathbf{x} } ^ \lambda) {\rm trace} (Z _\lambda   ) \nonumber\\
&+  {\rm trace} ( X ^\top R _\lambda X A_N  M _{E_\lambda |X} ^\top M _{E_\lambda |X} ) + {\rm trace}( \left(X ^\top  R_\lambda X A _N - \mathbb{I} _N \right) X ^\top 
B _\lambda M _{E_\lambda |X}  )\Big).
\end{align}
Using \eqref{proof trace ee x}, \eqref{proof trace mlml x}, and \eqref{proof E xmle x}  in   \eqref{proof mse training} yields the expression \eqref{MSE longer}. At the same time,   in \eqref{MSE longer} the last term can be rewritten as
\begin{align}
\label{third term shorter}
-\dfrac{2}{N} \  {\rm trace}\left( (X ^\top R _\lambda X A _N  -\mathbb{I}_N ) X^\top B _\lambda M _{E_\lambda |X}  \right) &= -\dfrac{2}{N} \  {\rm trace}\left( X ^\top ( Z _\lambda   -\mathbb{I}_p ) B _\lambda M _{E_\lambda |X}  \right) \nonumber\\
&= {2}\lambda{\rm trace}(   X ^\top R _\lambda  B _\lambda M _{E_\lambda |X} ) ,
\end{align}
by using  \eqref{Z1}. The substitution of this expression in
\eqref{MSE longer} provides the relation \eqref{MSE shorter}.
$\blacksquare$

\subsection{Proof of Corollary \ref{Corollary error Example A}}
\label{Proof of Corollary error Example A}

In order to prove \eqref{MSE training example A}, we just determine  all the  ingredients required for the evaluation of \eqref{MSE training} in the conditions of Example~A and we then show that  the result follows. First, notice that by \eqref{sigma lambda x}
\begin{align}
\label{sigma lambda x repeat}
\Sigma ^\lambda_{\boldsymbol{\varepsilon} | \mathbf{x} }&=  \Sigma _ { \boldsymbol{\varepsilon} }
\end{align}
and that by \eqref{mu eps lambda cond x example}
\begin{equation}
\label{mu eps lambda cond x example repeat}
M _{E_\lambda |X} = (B-B _\lambda ) ^\top X.
\end{equation}
Additionally,
\begin{align}
\label{}
M _{E_\lambda |X} ^\top M _{E_\lambda |X} = X ^\top (B-B _\lambda ) (B-B _\lambda ) ^\top X = X ^\top B B^\top X - X ^\top B B _\lambda ^\top X - X^\top B _\lambda B ^\top X + X ^\top B_\lambda B_\lambda ^\top X. 
\end{align}
We now proceed by pointing out that $M _{(\widehat{B}_\lambda - B _\lambda ) } = M _{\widehat{B}_\lambda  } + B _\lambda$ and by using that in the conditions of Example A, the mean matrix  $M _{\widehat{B}_\lambda  }$ is given by $M _{\widehat{B}_\lambda  } = Z _\lambda B$, with $Z _\lambda $ as in \eqref{Z lambda B lambda}. We hence have that 
\begin{align}M _{(\widehat{B}_\lambda - B _\lambda ) }&= B _\lambda + Z _\lambda B = B _\lambda + (\mathbb{I}_p - \lambda N R _\lambda ) B = B+B _\lambda - \lambda N R _\lambda B. \label{mu M example}
\end{align}
Additionally, we compute
\begin{align}
\label{mu M mu M t}
M _{(\widehat{B}_\lambda - B _\lambda ) }M _{(\widehat{B}_\lambda - B _\lambda ) } ^\top &= (B+B _\lambda - \lambda N R _\lambda B)(B+B _\lambda - \lambda N R _\lambda B)^\top =B B^\top - B B _\lambda^\top -  B_\lambda B ^\top + B _\lambda B _\lambda^\top          \nonumber\\
&-  \lambda N B B ^\top R _\lambda -  \lambda N R_\lambda  B B  ^\top +\lambda N R_\lambda  B B   _\lambda ^\top + \lambda N B _\lambda B ^\top R_\lambda + \lambda ^2 N ^2 R _\lambda B B ^\top R _\lambda.
\end{align}
We now substitute these expressions in the five summands that make formula \eqref{MSE shorter} and that we analyze one. First,
\begin{align}
\label{term1 example}
&{\rm trace} (\Sigma _{  \boldsymbol{\varepsilon}  | \mathbf{x} } ^ \lambda ) = {\rm trace} ( \Sigma _ { \boldsymbol{\varepsilon} } ),
\end{align}
\begin{align}
\label{term2 example}
\dfrac{1}{N} {\rm trace}(\Sigma _{  \boldsymbol{\varepsilon}  | \mathbf{x} } ^ \lambda) {\rm trace}\left( Z _\lambda  (R_\lambda X X ^\top - 2 \mathbb{I}_p )\right)= \dfrac{1}{N} {\rm trace}(\Sigma _ { \boldsymbol{\varepsilon} }) {\rm trace}\left( Z _\lambda  (R_\lambda X X ^\top - 2 \mathbb{I}_p )\right),\end{align}
\begin{align}
\label{term3 example}
\dfrac{1}{N} {\rm trace} \left( M _{(\widehat{B}_\lambda - B _\lambda ) }M _{(\widehat{B}_\lambda - B _\lambda ) }^\top X X ^\top \right) =& \dfrac{1}{N} {\rm trace} ( (B B^\top - B B _\lambda^\top - B_\lambda B ^\top + B _\lambda B _\lambda^\top - \lambda N B B ^\top R _\lambda      \nonumber\\ 
&- \lambda N R _\lambda B B ^\top+ \lambda N R_\lambda  B B _\lambda ^\top + \lambda N B _\lambda B ^\top R_\lambda \nonumber\\
& + \lambda ^2 N ^2 R _\lambda B B ^\top  _\lambda R _\lambda) X X ^\top ),\end{align}
\begin{align}
\label{term4 example}
 - \dfrac{2}{N}{\rm trace}((X ^\top R _\lambda X A_N  &- \dfrac{1}{2} \mathbb{I}_N ) M _{E_\lambda |X} ^\top M _{E_\lambda |X}) =  - \dfrac{2}{N}{\rm trace}((X ^\top R _\lambda X A_N  - \dfrac{1}{2} \mathbb{I}_N ) (X ^\top B B^\top X - X ^\top B B _\lambda ^\top X\nonumber\\
& - X^\top B _\lambda B ^\top X + X ^\top B_\lambda B_\lambda ^\top X)) = \dfrac{1}{N}{\rm trace} (( - B B ^\top + B B _\lambda ^\top + B _\lambda B ^\top - B _\lambda B _\lambda ^\top \nonumber\\
&+ 2 \lambda N  R_\lambda B B ^\top - 2 \lambda N R _\lambda B B _\lambda ^\top  - 2\lambda N R _\lambda B _\lambda B ^\top + 2 \lambda N  R _\lambda B _\lambda B _\lambda ^\top )X X^\top ),\end{align}
and finally,
\begin{align}\label{term5 example}
{2}\lambda{\rm trace}\left(   X ^\top R _\lambda  B _\lambda M _{E_\lambda |X} \right) = {2}\lambda{\rm trace}\left(   X ^\top R _\lambda  B _\lambda  (B-B _\lambda ) ^\top X \right) = {2}\lambda{\rm trace}\left(    (R _\lambda  B _\lambda  B ^\top -R _\lambda  B _\lambda B _\lambda^\top )  X X ^\top\right),
\end{align}
respectively.

Gathering the terms \eqref{term1 example}-\eqref{term5 example}, we obtain that
\begin{align*}
{\rm MSE}_{{\rm training}|X}^\lambda = &{\rm trace} ( \Sigma _ { \boldsymbol{\varepsilon} } ) + \dfrac{1}{N} {\rm trace}(\Sigma _ { \boldsymbol{\varepsilon} }) {\rm trace}\left( Z _\lambda  (R_\lambda X X ^\top - 2 \mathbb{I}_p )\right) \nonumber\\
&+ \dfrac{1}{N} {\rm trace}( (B B^\top - B B _\lambda^\top - B_\lambda B ^\top + B _\lambda B _\lambda^\top - \lambda N B B ^\top R _\lambda  - \lambda N R_\lambda  B B  ^\top   \nonumber\\ 
&+ \lambda N B _\lambda B ^\top R_\lambda  + \lambda N R_\lambda B  B _\lambda^\top  + \lambda ^2 N ^2 R _\lambda B B ^\top   R_\lambda - B B ^\top + B B _\lambda ^\top + B _\lambda B ^\top - B _\lambda B _\lambda ^\top\nonumber\\
&+ 2 \lambda N R _\lambda B B ^\top - 2 \lambda N R _\lambda B B _\lambda ^\top - 2 \lambda N R _\lambda B _\lambda B ^\top + 2 \lambda N  R _\lambda B _\lambda B _\lambda ^\top + 2 \lambda N   R _\lambda  B _\lambda B ^\top        \nonumber\\&- 2\lambda N R _\lambda B _\lambda B_ \lambda  ^\top   ) X X ^\top)\nonumber\\
	&= {\rm trace} ( \Sigma _ { \boldsymbol{\varepsilon} } ) + \dfrac{1}{N} {\rm trace}(\Sigma _ { \boldsymbol{\varepsilon} }) {\rm trace}\left( Z _\lambda  (R_\lambda X X ^\top - 2 \mathbb{I}_p )\right) +  \lambda ^2 N {\rm trace}( R_\lambda B B ^\top   R_\lambda X X ^\top),
\end{align*} 
which coincides with \eqref{MSE training example A}, as required. $\blacksquare$  

\subsection{Proof of Theorem \ref{Theorem testing error}}
\label{Proof of Theorem testing error}
In order to prove \eqref{MSE test}, we first rewrite \eqref{MSE testing} as
\begin{align}
\label{MSE test proof}
{\rm MSE} ^\lambda _{{\rm testing}|X _1 } &= \dfrac{1}{N _2 } {\rm E}_{X _1} \left[ \left(Y _2 - \widehat {B} _\lambda ^\top X _2  \right) ^\top \left(Y _2 - \widehat {B} _\lambda ^\top X _2  \right)   \right] \nonumber\\
&=\dfrac{1}{N _2 } \left[ {\rm trace}  \left( {\rm E}_{X _1} \left[ Y _2 ^\top Y _2   \right] \right) - 2\, {\rm trace}  \left( {\rm E}_{X _1} \left[   Y_2^\top \widehat{B} _\lambda ^\top X _2  \right] \right) + {\rm trace} \left( {\rm E} _{X _1}\left[ X _2 ^\top \widehat{B} _\lambda \widehat{B} _\lambda^\top X _2   \right] \right) \right].
\end{align}
In order to evaluate this expression, we study separately all  its terms. We start with the first summand and use that the pairs of random samples  $(X_1, Y_1)$ and $(X_2, Y_2)$ used for training and testing, respectively, are independent from each other. We hence rewrite the first term in \eqref{MSE test proof} as
\begin{align}
\label{term1}
\dfrac{1}{N _2 }  {\rm trace}  \left( {\rm E} _{X _1}\left[ Y _2 ^\top Y _2   \right] \right) = \dfrac{1}{N _2 }  {\rm trace}  \left( {\rm E} \left[ Y _2 ^\top Y _2   \right] \right) = {\rm trace} \left[ \Sigma  ^{(2)} _{\mathbf{y} } +\boldsymbol{\mu} ^{(2)} _{\mathbf{y} } \boldsymbol{\mu} _{\mathbf{y}  }^ {(2)\top}  \right]
\end{align}
with \begin{align}
\label{}
\Sigma  ^{(2)} _{\mathbf{y} }   &= {\rm Cov} ( \mathbf{y} ^{(2)} ,\mathbf{y} ^{(2)} ),\\
\boldsymbol{\mu} ^{(2)} _{\mathbf{y} } &= {\rm E} \left[ \mathbf{y} ^{(2)}\right].
\end{align}
For the second summand we obtain
 \begin{align}
\label{term2}
- \dfrac{2}{N_2} {\rm trace}  \left( {\rm E}_{X _1} \left[   Y_2^\top \widehat{B} _\lambda ^\top X _2  \right] \right) &= - \dfrac{2}{N_2} {\rm trace}  \left( {\rm E}_{X _1} \left[   X _2 Y_2^\top \widehat{B} _\lambda ^\top \right] \right) = - \dfrac{2}{N_2} {\rm trace}  \left( {\rm E} \left[   X _2 Y_2^\top\right] {\rm E} _{X _1}\left[  \widehat{B} _\lambda ^\top  \right] \right)\nonumber\\
 &=- {2}{} {\rm trace}  \left( \left( \Sigma  ^{(2)}_{\mathbf{x}   \mathbf{y}  }  + \boldsymbol{\mu} _{ \mathbf{x}   }^{(2)}\boldsymbol{\mu} _{\mathbf{y} } ^{(2)\top}\right) M ^{  \top}_{ \widehat{B} _\lambda}\right),
\end{align}
with 
\begin{align}
\label{mu B hat}
M _{ \widehat{B} _\lambda}&:=B _\lambda   -\lambda N _1 R _\lambda B _\lambda   + R _\lambda X _1 A _{N_1} M _{ E_\lambda | X} ^{(1) \top},\\
\Sigma  ^{(2)} _{\mathbf{x}\mathbf{y} }   &= {\rm Cov} ( \mathbf{x} ^{(2)} ,\mathbf{y} ^{(2)} ),\\
\boldsymbol{\mu} ^{(2)} _{\mathbf{x} } &= {\rm E} \left[ \mathbf{x} ^{(2)}\right].
\end{align}
In \eqref{term2} we used the properties of the ridge regression matrix estimator $\widehat{B} _\lambda$ provided in Theorem~\ref{Theorem 1}. Finally, we rewrite the third term as
 \begin{align}
\label{term3}
\dfrac{1}{N_2}{\rm trace} \left( {\rm E}_{X _1} \left[ X _2 ^\top \widehat{B} _\lambda \widehat{B} _\lambda^\top X _2   \right] \right)&= \dfrac{1}{N_2}{\rm trace} \left( {\rm E} _{X _1}\left[  X _2 X _2 ^\top \widehat{B} _\lambda \widehat{B} _\lambda^\top  \right] \right) =\dfrac{1}{N_2}{\rm trace} \left( {\rm E} \left[  X _2 X _2 ^\top \right]  {\rm E} _{X _1}\left[ \widehat{B} _\lambda \widehat{B} _\lambda^\top \right] \right) \nonumber\\
 &={\rm trace} \left[ \left( \Sigma^{(2)}  _{\mathbf{x}  }  + \boldsymbol{\mu}^{(2)} _{\mathbf{x}  } \boldsymbol{\mu} _{\mathbf{x}  }^ {(2)\top} \right) \left({\rm trace} \left(\Sigma ^{\lambda, (1)}_{ \boldsymbol{\varepsilon} | \mathbf{x} }\right)  Z _\lambda   R _\lambda   + M_{ \widehat{B} _\lambda}M ^{  \top}_{ \widehat{B} _\lambda} \right)\right],
\end{align}
with 
\begin{align}
\label{}
\Sigma  ^{(2)} _{\mathbf{x} }   &= {\rm Cov} ( \mathbf{x} ^{(2)} ,\mathbf{x} ^{(2)} ),
\end{align}
and where $M _{ \widehat{B} _\lambda}$ is given in \eqref{mu B hat}. In this expression we again used Theorem~\ref{Theorem 1} and the property \eqref{covRowproperty}.
Substituting expressions \eqref{term1}-\eqref{term3} into \eqref{MSE test proof} immediately yields \eqref{MSE test}, as required. $\blacksquare$

\noindent
\addcontentsline{toc}{section}{Bibliography}
\bibliographystyle{wmaainf}
\bibliography{/Users/JP17/Dropbox/Public/GOLibrary}
\end{document}